\documentclass[a4paper,10pt]{article}

\usepackage[margin=3cm]{geometry}
\usepackage[numbers]{natbib}
\usepackage{hyperref}
\usepackage{booktabs}
\usepackage{verbatim}
\usepackage{url}            
\usepackage{graphicx}
\usepackage[symbol]{footmisc}

\usepackage{natbib,float}
\usepackage{algorithm,algorithmic}

\usepackage{amsfonts,amsmath,amssymb,amsthm}

\usepackage{dirtree}

\usepackage{xcolor}
\usepackage{tcolorbox}
\tcbuselibrary{listings}
\definecolor{light-gray}{gray}{0.95}
\usepackage{listings}

\DeclareMathOperator*{\argmax}{arg\,max}
\DeclareMathOperator*{\argmin}{arg\,min}

\usepackage{mathpazo}

\usepackage[labelfont=bf]{caption}
\usepackage{subcaption}

\usepackage{adjustbox} 
\usepackage{enumerate} 
\usepackage{textcomp} 
\AtBeginDocument{%
}
\usepackage{upquote} 
\usepackage[mathletters]{ucs} 
\usepackage[utf8x]{inputenc} 
\usepackage{fancyvrb} 
\usepackage{grffile} 
\usepackage{longtable} 
\usepackage[inline]{enumitem} 
\usepackage[normalem]{ulem} 

\definecolor{urlcolor}{rgb}{0,.145,.698}
\definecolor{linkcolor}{rgb}{.71,0.21,0.01}
\definecolor{citecolor}{rgb}{.12,.54,.11}

\definecolor{ansi-black}{HTML}{3E424D}
\definecolor{ansi-black-intense}{HTML}{282C36}
\definecolor{ansi-red}{HTML}{E75C58}
\definecolor{ansi-red-intense}{HTML}{B22B31}
\definecolor{ansi-green}{HTML}{00A250}
\definecolor{ansi-green-intense}{HTML}{007427}
\definecolor{ansi-yellow}{HTML}{DDB62B}
\definecolor{ansi-yellow-intense}{HTML}{B27D12}
\definecolor{ansi-blue}{HTML}{208FFB}
\definecolor{ansi-blue-intense}{HTML}{0065CA}
\definecolor{ansi-magenta}{HTML}{D160C4}
\definecolor{ansi-magenta-intense}{HTML}{A03196}
\definecolor{ansi-cyan}{HTML}{60C6C8}
\definecolor{ansi-cyan-intense}{HTML}{258F8F}
\definecolor{ansi-white}{HTML}{C5C1B4}
\definecolor{ansi-white-intense}{HTML}{A1A6B2}

\DefineVerbatimEnvironment{Highlighting}{Verbatim}{commandchars=\\\{\}}

\title{test}

\makeatletter
\def\PY@reset{\let\PY@it=\relax \let\PY@bf=\relax%
\let\PY@ul=\relax \let\PY@tc=\relax%
\let\PY@bc=\relax \let\PY@ff=\relax}
\def\PY@tok#1{\csname PY@tok@#1\endcsname}
\def\PY@toks#1+{\ifx\relax#1\empty\else%
\PY@tok{#1}\expandafter\PY@toks\fi}
\def\PY@do#1{\PY@bc{\PY@tc{\PY@ul{%
\PY@it{\PY@bf{\PY@ff{#1}}}}}}}
\def\PY#1#2{\PY@reset\PY@toks#1+\relax+\PY@do{#2}}

\expandafter\def\csname PY@tok@w\endcsname{\def\PY@tc##1{\textcolor[rgb]{0.73,0.73,0.73}{##1}}}
\expandafter\def\csname PY@tok@c\endcsname{\let\PY@it=\textit\def\PY@tc##1{\textcolor[rgb]{0.25,0.50,0.50}{##1}}}
\expandafter\def\csname PY@tok@cp\endcsname{\def\PY@tc##1{\textcolor[rgb]{0.74,0.48,0.00}{##1}}}
\expandafter\def\csname PY@tok@k\endcsname{\let\PY@bf=\textbf\def\PY@tc##1{\textcolor[rgb]{0.00,0.50,0.00}{##1}}}
\expandafter\def\csname PY@tok@kp\endcsname{\def\PY@tc##1{\textcolor[rgb]{0.00,0.50,0.00}{##1}}}
\expandafter\def\csname PY@tok@kt\endcsname{\def\PY@tc##1{\textcolor[rgb]{0.69,0.00,0.25}{##1}}}
\expandafter\def\csname PY@tok@o\endcsname{\def\PY@tc##1{\textcolor[rgb]{0.40,0.40,0.40}{##1}}}
\expandafter\def\csname PY@tok@ow\endcsname{\let\PY@bf=\textbf\def\PY@tc##1{\textcolor[rgb]{0.67,0.13,1.00}{##1}}}
\expandafter\def\csname PY@tok@nb\endcsname{\def\PY@tc##1{\textcolor[rgb]{0.00,0.50,0.00}{##1}}}
\expandafter\def\csname PY@tok@nf\endcsname{\def\PY@tc##1{\textcolor[rgb]{0.00,0.00,1.00}{##1}}}
\expandafter\def\csname PY@tok@nc\endcsname{\let\PY@bf=\textbf\def\PY@tc##1{\textcolor[rgb]{0.00,0.00,1.00}{##1}}}
\expandafter\def\csname PY@tok@nn\endcsname{\let\PY@bf=\textbf\def\PY@tc##1{\textcolor[rgb]{0.00,0.00,1.00}{##1}}}
\expandafter\def\csname PY@tok@ne\endcsname{\let\PY@bf=\textbf\def\PY@tc##1{\textcolor[rgb]{0.82,0.25,0.23}{##1}}}
\expandafter\def\csname PY@tok@nv\endcsname{\def\PY@tc##1{\textcolor[rgb]{0.10,0.09,0.49}{##1}}}
\expandafter\def\csname PY@tok@no\endcsname{\def\PY@tc##1{\textcolor[rgb]{0.53,0.00,0.00}{##1}}}
\expandafter\def\csname PY@tok@nl\endcsname{\def\PY@tc##1{\textcolor[rgb]{0.63,0.63,0.00}{##1}}}
\expandafter\def\csname PY@tok@ni\endcsname{\let\PY@bf=\textbf\def\PY@tc##1{\textcolor[rgb]{0.60,0.60,0.60}{##1}}}
\expandafter\def\csname PY@tok@na\endcsname{\def\PY@tc##1{\textcolor[rgb]{0.49,0.56,0.16}{##1}}}
\expandafter\def\csname PY@tok@nt\endcsname{\let\PY@bf=\textbf\def\PY@tc##1{\textcolor[rgb]{0.00,0.50,0.00}{##1}}}
\expandafter\def\csname PY@tok@nd\endcsname{\def\PY@tc##1{\textcolor[rgb]{0.67,0.13,1.00}{##1}}}
\expandafter\def\csname PY@tok@s\endcsname{\def\PY@tc##1{\textcolor[rgb]{0.73,0.13,0.13}{##1}}}
\expandafter\def\csname PY@tok@sd\endcsname{\let\PY@it=\textit\def\PY@tc##1{\textcolor[rgb]{0.73,0.13,0.13}{##1}}}
\expandafter\def\csname PY@tok@si\endcsname{\let\PY@bf=\textbf\def\PY@tc##1{\textcolor[rgb]{0.73,0.40,0.53}{##1}}}
\expandafter\def\csname PY@tok@se\endcsname{\let\PY@bf=\textbf\def\PY@tc##1{\textcolor[rgb]{0.73,0.40,0.13}{##1}}}
\expandafter\def\csname PY@tok@sr\endcsname{\def\PY@tc##1{\textcolor[rgb]{0.73,0.40,0.53}{##1}}}
\expandafter\def\csname PY@tok@ss\endcsname{\def\PY@tc##1{\textcolor[rgb]{0.10,0.09,0.49}{##1}}}
\expandafter\def\csname PY@tok@sx\endcsname{\def\PY@tc##1{\textcolor[rgb]{0.00,0.50,0.00}{##1}}}
\expandafter\def\csname PY@tok@m\endcsname{\def\PY@tc##1{\textcolor[rgb]{0.40,0.40,0.40}{##1}}}
\expandafter\def\csname PY@tok@gh\endcsname{\let\PY@bf=\textbf\def\PY@tc##1{\textcolor[rgb]{0.00,0.00,0.50}{##1}}}
\expandafter\def\csname PY@tok@gu\endcsname{\let\PY@bf=\textbf\def\PY@tc##1{\textcolor[rgb]{0.50,0.00,0.50}{##1}}}
\expandafter\def\csname PY@tok@gd\endcsname{\def\PY@tc##1{\textcolor[rgb]{0.63,0.00,0.00}{##1}}}
\expandafter\def\csname PY@tok@gi\endcsname{\def\PY@tc##1{\textcolor[rgb]{0.00,0.63,0.00}{##1}}}
\expandafter\def\csname PY@tok@gr\endcsname{\def\PY@tc##1{\textcolor[rgb]{1.00,0.00,0.00}{##1}}}
\expandafter\def\csname PY@tok@ge\endcsname{\let\PY@it=\textit}
\expandafter\def\csname PY@tok@gs\endcsname{\let\PY@bf=\textbf}
\expandafter\def\csname PY@tok@gp\endcsname{\let\PY@bf=\textbf\def\PY@tc##1{\textcolor[rgb]{0.00,0.00,0.50}{##1}}}
\expandafter\def\csname PY@tok@go\endcsname{\def\PY@tc##1{\textcolor[rgb]{0.53,0.53,0.53}{##1}}}
\expandafter\def\csname PY@tok@gt\endcsname{\def\PY@tc##1{\textcolor[rgb]{0.00,0.27,0.87}{##1}}}
\expandafter\def\csname PY@tok@err\endcsname{\def\PY@bc##1{\setlength{\fboxsep}{0pt}\fcolorbox[rgb]{1.00,0.00,0.00}{1,1,1}{\strut ##1}}}
\expandafter\def\csname PY@tok@kc\endcsname{\let\PY@bf=\textbf\def\PY@tc##1{\textcolor[rgb]{0.00,0.50,0.00}{##1}}}
\expandafter\def\csname PY@tok@kd\endcsname{\let\PY@bf=\textbf\def\PY@tc##1{\textcolor[rgb]{0.00,0.50,0.00}{##1}}}
\expandafter\def\csname PY@tok@kn\endcsname{\let\PY@bf=\textbf\def\PY@tc##1{\textcolor[rgb]{0.00,0.50,0.00}{##1}}}
\expandafter\def\csname PY@tok@kr\endcsname{\let\PY@bf=\textbf\def\PY@tc##1{\textcolor[rgb]{0.00,0.50,0.00}{##1}}}
\expandafter\def\csname PY@tok@bp\endcsname{\def\PY@tc##1{\textcolor[rgb]{0.00,0.50,0.00}{##1}}}
\expandafter\def\csname PY@tok@fm\endcsname{\def\PY@tc##1{\textcolor[rgb]{0.00,0.00,1.00}{##1}}}
\expandafter\def\csname PY@tok@vc\endcsname{\def\PY@tc##1{\textcolor[rgb]{0.10,0.09,0.49}{##1}}}
\expandafter\def\csname PY@tok@vg\endcsname{\def\PY@tc##1{\textcolor[rgb]{0.10,0.09,0.49}{##1}}}
\expandafter\def\csname PY@tok@vi\endcsname{\def\PY@tc##1{\textcolor[rgb]{0.10,0.09,0.49}{##1}}}
\expandafter\def\csname PY@tok@vm\endcsname{\def\PY@tc##1{\textcolor[rgb]{0.10,0.09,0.49}{##1}}}
\expandafter\def\csname PY@tok@sa\endcsname{\def\PY@tc##1{\textcolor[rgb]{0.73,0.13,0.13}{##1}}}
\expandafter\def\csname PY@tok@sb\endcsname{\def\PY@tc##1{\textcolor[rgb]{0.73,0.13,0.13}{##1}}}
\expandafter\def\csname PY@tok@sc\endcsname{\def\PY@tc##1{\textcolor[rgb]{0.73,0.13,0.13}{##1}}}
\expandafter\def\csname PY@tok@dl\endcsname{\def\PY@tc##1{\textcolor[rgb]{0.73,0.13,0.13}{##1}}}
\expandafter\def\csname PY@tok@s2\endcsname{\def\PY@tc##1{\textcolor[rgb]{0.73,0.13,0.13}{##1}}}
\expandafter\def\csname PY@tok@sh\endcsname{\def\PY@tc##1{\textcolor[rgb]{0.73,0.13,0.13}{##1}}}
\expandafter\def\csname PY@tok@s1\endcsname{\def\PY@tc##1{\textcolor[rgb]{0.73,0.13,0.13}{##1}}}
\expandafter\def\csname PY@tok@mb\endcsname{\def\PY@tc##1{\textcolor[rgb]{0.40,0.40,0.40}{##1}}}
\expandafter\def\csname PY@tok@mf\endcsname{\def\PY@tc##1{\textcolor[rgb]{0.40,0.40,0.40}{##1}}}
\expandafter\def\csname PY@tok@mh\endcsname{\def\PY@tc##1{\textcolor[rgb]{0.40,0.40,0.40}{##1}}}
\expandafter\def\csname PY@tok@mi\endcsname{\def\PY@tc##1{\textcolor[rgb]{0.40,0.40,0.40}{##1}}}
\expandafter\def\csname PY@tok@il\endcsname{\def\PY@tc##1{\textcolor[rgb]{0.40,0.40,0.40}{##1}}}
\expandafter\def\csname PY@tok@mo\endcsname{\def\PY@tc##1{\textcolor[rgb]{0.40,0.40,0.40}{##1}}}
\expandafter\def\csname PY@tok@ch\endcsname{\let\PY@it=\textit\def\PY@tc##1{\textcolor[rgb]{0.25,0.50,0.50}{##1}}}
\expandafter\def\csname PY@tok@cm\endcsname{\let\PY@it=\textit\def\PY@tc##1{\textcolor[rgb]{0.25,0.50,0.50}{##1}}}
\expandafter\def\csname PY@tok@cpf\endcsname{\let\PY@it=\textit\def\PY@tc##1{\textcolor[rgb]{0.25,0.50,0.50}{##1}}}
\expandafter\def\csname PY@tok@c1\endcsname{\let\PY@it=\textit\def\PY@tc##1{\textcolor[rgb]{0.25,0.50,0.50}{##1}}}
\expandafter\def\csname PY@tok@cs\endcsname{\let\PY@it=\textit\def\PY@tc##1{\textcolor[rgb]{0.25,0.50,0.50}{##1}}}

\makeatother

\definecolor{incolor}{rgb}{0.0, 0.0, 0.5}
\definecolor{outcolor}{rgb}{0.545, 0.0, 0.0}

\hypersetup{
  breaklinks=true,  
  colorlinks=true,
  urlcolor=urlcolor,
  linkcolor=linkcolor,
  citecolor=citecolor,
  }

\newcommand{\bbox}[2]{\fbox{\begin{minipage}{\textwidth}
  \texttt{#1\hfill #2}
\end{minipage}}\vspace{2mm}}

\title{{\tt Adversarial Robustness Toolbox} v1.0.0}
\date{}
\author{Maria-Irina Nicolae\footnote{maria-irina.nicolae@ibm.com} $^{,1}$, Mathieu Sinn\footnote{mathsinn@ie.ibm.com} $^{,1}$, Minh Ngoc Tran$^1$, Beat Buesser$^1$, \\ Ambrish Rawat$^1$,  Martin Wistuba$^1$, Valentina Zantedeschi\footnote{Contributed to ART while doing an internship with IBM Research -- Ireland.} $^{,2}$, Nathalie Baracaldo$^3$, \\ Bryant Chen$^3$, Heiko Ludwig$^3$, Ian M.~Molloy$^4$, Ben Edwards$^4$ \\ \small $^1$IBM Research -- Ireland \\ \small $^2$Univ Lyon, UJM-Saint-Etienne, CNRS, Institut d Optique Graduate School, \\ \small Laboratoire Hubert Curien UMR 5516, France \\ \small $^3$IBM Research -- Almaden \\ \small $^4$IBM Research -- Yorktown Heights}

\begin{document}

\maketitle

\begin{abstract}

Adversarial Robustness Toolbox (ART) is a Python library supporting developers and researchers in defending Machine Learning models (Deep Neural Networks, Gradient Boosted Decision Trees, Support Vector Machines, Random Forests, Logistic Regression, Gaussian Processes, Decision Trees, Scikit-learn Pipelines, etc.) against adversarial threats and helps making AI systems more secure and trustworthy. Machine Learning models are vulnerable to adversarial examples, which are inputs (images, texts, tabular data, etc.) deliberately modified to produce a desired response by the Machine Learning model. ART provides the tools to build and deploy defences and test them with adversarial attacks.

Defending Machine Learning models involves certifying and verifying model robustness and model hardening with approaches such as pre-processing inputs, augmenting training data with adversarial samples, and leveraging runtime detection methods to flag any inputs that might have been modified by an adversary. The attacks implemented in ART allow creating adversarial attacks against Machine Learning models which is required to test defenses with state-of-the-art threat models.

Supported Machine Learning Libraries include TensorFlow (v1 and v2), Keras, PyTorch, MXNet, Scikit-learn, XGBoost, LightGBM, CatBoost, and GPy.

The source code of ART is released with MIT license at \url{https://github.com/IBM/adversarial-robustness-toolbox}. The release includes code examples, notebooks with tutorials and documentation (\url{http://adversarial-robustness-toolbox.readthedocs.io}).
\end{abstract}

\newpage

\color{black} 
\tableofcontents

\newpage



\section{Introduction}\label{sec:introduction}

The Adversarial Robustness Toolbox (ART) is an open-source Python library for adversarial machine learning. It has been released under an MIT license and is available at \url{https://github.com/IBM/adversarial-robustness-toolbox}. It provides standardized interfaces for classifiers of the most popular machine learning libraries (TensorFlow, Keras, PyTorch, MXNet, Scikit-learn, XGBoost, LightGBM, CatBoost, and GPy. The architecture of ART makes it easy to combine defences, e.g.\ adversarial training with data preprocessing and runtime detection of adversarial inputs. ART is designed both for researchers who want to run large-scale experiments for benchmarking novel attacks or defences, and for developers composing and deploying secure machine learning applications.

Here we provide mathematical background and implementation details for ART. It is complementary to the documentation hosted on Read the Docs (\url{http://adversarial-robustness-toolbox.readthedocs.io}). In particular, it explains the semantics and mathematical backgrounds of attacks and defences, and highlights custom choices in the implementation.

This document is structured as follows: 

Section~\ref{sec:background} provides background and introduces mathematical notation.
Section~\ref{sec:library_modules} gives an overview of the ART architecture and library modules.

The following sections cover the different modules in detail: Section~\ref{sec:classifier} introduces the classifier modules, Section~\ref{sec:attacks} the evasion attacks and Section~\ref{sec:defences} the evasion defences.
Section~\ref{sec:detection} covers detection of evasion attacks, while Section~\ref{sec:poison} detection for poisoning, and Section~\ref{sec:metrics} the metrics.
Section~\ref{sec:data_gen} details the wrappers for data generators.
Finally, Section~\ref{sec:versioning} describes the versioning system.

\section{Background} \label{sec:background}

While early work in machine learning has often assumed a closed and trusted environment, attacks against the machine learning process and resulting models have received increased attention in the past years.
\textbf{Adversarial machine learning} aims to protect the machine learning pipeline to ensure its safety at training, test and inference time \cite{DBLP:journals/corr/PapernotMSW16,barreno2010, biggio2013pattern}.

The threat of evasion attacks against machine learning models at test time was first highlighted by~\citep{biggio2013evasion}.
\citep{szegedy2013} investigated specifically the vulnerability of deep neural network (DNN) models and proposed an efficient algorithm for crafting adversarial examples for such models. Since then, there has been an explosion of work on proposing more advanced adversarial attacks, on understanding the phenomenon of adversarial examples, on assessing the robustness of specific DNN architectures and learning paradigms, and on proposing as well as evaluating various defence strategies against evasion attacks.

Generally, the objective of an {\bf evasion attack} is to modify the input to a {\bf classifier} such that it is misclassified, while keeping the modification as small as possible.
An important distinction is between {\bf untargeted} and {\bf targeted} attacks: If untargeted, the attacker aims for a misclassification of the modified input without any constraints on what the new class should be; if targeted, the new class is specified by the attacker.
Another important distinction is between {\bf black-box} and {\bf white-box} attacks: in the white-box case, the attacker has full access to the architecture and parameters of the classifier.
For a black-box attack, this is not the case.
A typical strategy there is to use a {\bf surrogate model} for crafting the attacks, and exploiting the {\bf transferability} of adversarial examples that has been demonstrated among a variety of architectures (in the image classification domain, at least).
Another way to approach the black-box threat model is through the use of zero-order optimization: attacks in this category are able to produce adversarial samples without accessing model gradients at all (e.g.\ ZOO~\citep{chen2017zoo}).
They rely on zero-order approximations of the target model.
One can also consider a wide range of {\bf grey-box} settings in which the attacker may not have access to the classifier's parameters, but to its architecture, training algorithm or training data.

On the {\bf adversarial defence} side, two different strategies can be considered: {\bf model hardening} and {\bf runtime detection} of adversarial inputs. Among the model hardening methods, a widely explored approach is to augment the training data of the classifier, e.g.\ by adversarial examples (so-called {\bf adversarial training}~\citep{goodfellow2014fgsm,miyato2017virtual}) or other augmentation methods. Another approach is {\bf input data preprocessing}, often using non-differentiable or randomized transformation~\citep{guo2018}, transformations reducing the dimensionality of the inputs~\citep{xu2017feature_squeeze}, or transformations aiming to project inputs onto the ``true'' data manifold~\citep{meng2017}.
Other model hardening approaches involve special types of {\bf regularization} during model training~\citep{ross2017}, or modifying elements of the classifier's architecture~\citep{zantedeschi2017}.

Note that {\bf robustness metrics} are a key element to measure the vulnerability of a classifier with respect to particular attacks, and to assess the effectiveness of adversarial defences. Typically such metrics quantify the amount of perturbation that is required to cause a misclassification or, more generally, the sensitivity of model outputs with respect to changes in their inputs.

\textbf{Poisoning attacks} are another threat to machine learning systems executed at data collection and training time. Machine learning systems often assume that the data used for training can be trusted and fully reflects the population of interest. However, data collection and curation processes are often not fully controlled by the owner or stakeholders of the model. For example, common data sources include social media, crowdsourcing, consumer behavior and internet of the things measurements. This lack of control creates a threat of poisoning attacks where adversaries have the opportunity of manipulating the training data to significantly decrease overall performance, cause targeted misclassification or bad behavior, and insert backdoors and neural trojans \cite{barreno2010,ch-roni-nelson2010,huang2011adv,badnets,liu2017trojaning,liu2017neural,biggio2013poisoning,munoz2017}.
Defences for this threat aim to detect and filter malicious training data~\cite{barreno2010,Baracaldo:2017,DBLP:journals/corr/PapernotMSW16}.

\paragraph{Mathematical notation}
In the remainder of this section, we are introducing mathematical notation that will be used for the explanation of the various attacks and defence techniques in the following. Table~\ref{table:notation} lists the key notation for quick reference.


\begin{table}[h]
\begin{center}
\begin{tabular}{ll}
\toprule
Notation & Description \\
\midrule
${\cal X}$ & Space of classifier inputs \\
$\boldsymbol{x}$ & Classifier input \\
$X$ & Random variable \\
$x_{\mbox{\tiny min}}$ & Minimum clipping value for classifier inputs \\
$x_{\mbox{\tiny max}}$ & Maximum clipping value for classifier inputs \\
$\mbox{clip}(\boldsymbol{x}, x_{\mbox{\tiny min}}, x_{\mbox{\tiny max}})$ & Function clipping $\boldsymbol{x}$ at $x_{\mbox{\tiny min}}$ and $x_{\mbox{\tiny max}}$, respectively \\
$\|\boldsymbol{x}\|_p$ & $\ell_p$ norm of $\boldsymbol{x}$ \\
$\mbox{project}(\boldsymbol{x}, p, \epsilon)$ & Function returning $\boldsymbol{x}'$ with smallest norm $\|\boldsymbol{x}-\boldsymbol{x}'\|_2$ satisfying $\|\boldsymbol{x}'\|_p\leq \epsilon$\\
${\cal Y}$ & Space of classifier outputs (=labels) \\
$K$ & Cardinality of ${\cal Y}$ (note: we assume ${\cal Y}=\{1,\ldots,K\}$) \\
$y$ & Label \\
$Z(\boldsymbol{x})$ & Classifier logits (ranging in ${\mathbb R}^K$) \\
$F(\boldsymbol{x})$ & Class probabilities ($=\mbox{softmax}(Z(\boldsymbol{x}))$)\\
$C(\boldsymbol{x})$ & Classification of input $\boldsymbol{x}$ (also used to denote the classifier itself) \\
$\rho(\boldsymbol{x})$ & Untargeted adversarial perturbation \\
$\psi(\boldsymbol{x}, y)$ & Targeted adversarial perturbation \\
$\boldsymbol{x}_{\mbox{\tiny adv}}$ & Adversarial sample \\
${\cal L}(\boldsymbol{x}, y)$ & Loss function \\
$\nabla_{\boldsymbol{x}}{\cal L}(\boldsymbol{x}, y)$ & Loss gradients \\
$\nabla Z(\boldsymbol{x})$ & Logit gradients \\
$\nabla F(\boldsymbol{x})$ & Output gradients \\
$\boldsymbol{x} \odot \boldsymbol{z}$ & Element-wise product \\
\bottomrule
\end{tabular}
\end{center}
\caption{Summary of notation.}
\label{table:notation}
\end{table}

The notion of {\bf classifier} will be central in the following. By $\boldsymbol{x}\in{\cal X}$ we denote the {\bf classifier inputs}. For most parts, we are concerned with classifier inputs that are images and hence assume the {\bf space of classifier inputs} is  ${\cal X}\subset{\mathbb R}^{k_1\times k_2 \times k_3}$, where $k_1$ is the width, $k_2$ the height, and $k_3$ the number of color channels of the image\footnote[4]{Typically $k_3=1$ or $k_3=3$ depending on whether $\boldsymbol{x}$ is a greyscale or colored image.}.
We assume that the classifier inputs have minimum and maximum {\bf clipping values} $x_{\mbox{\tiny min}}$ and
$x_{\mbox{\tiny max}}$, respectively, i.e.~each component of $\boldsymbol{x}$ lies within the interval $[x_{\mbox{\tiny min}},x_{\mbox{\tiny max}}]$.
For images, this interval is typically $[0,255]$ in case of $8$-bit pixel values, or $[0,1]$ if the pixel values have been normalized. For an arbitrary $\boldsymbol{x}\in {\cal X}$, we use $\mbox{clip}(\boldsymbol{x}, x_{\mbox{\tiny min}}, x_{\mbox{\tiny max}})$ to denote the input that is obtained by clipping each component of $\boldsymbol{x}$ at $x_{\mbox{\tiny min}}$ and $x_{\mbox{\tiny max}}$, respectively.
Moreover we write $\|\boldsymbol{x}\|_p$ for the $\ell_p$ {\bf norm} of $\boldsymbol{x}$, and
$\mbox{project}(\boldsymbol{x}, p, \epsilon)$ for the function which returns among all $\boldsymbol{x}'$ satisfying $\|\boldsymbol{x}'\|_p\leq \epsilon$ the one with smallest norm $\|\boldsymbol{x}-\boldsymbol{x}'\|_2$.
In the special case $p=2$ this is equivalent to multiplying $\boldsymbol{x}$ component-wise with the minimum of $1$ and $\epsilon/\|\boldsymbol{x}\|_2$, and in the case $p=\infty$ equivalent to clipping each component at $\pm\epsilon$.

By $y\in{\cal Y}$ we denote the {\bf classifier outputs}, which we also refer to as {\bf labels}. We always assume the {\bf space of classifier outputs} is
${\cal Y}=\{1,\ldots,K\}$, i.e.~there are $K$ different classes.
For most parts, we will consider classifiers based on a {\bf logit} function $Z: {\cal X} \to {\mathbb R}^K$. We refer to $Z(\boldsymbol{x})$ as the {\bf logits} of the classifier for input $\boldsymbol{x}$. The {\bf class probabilities} are obtained by applying the {\bf softmax} function, $F(\boldsymbol{x}) = \mbox{softmax}(Z(\boldsymbol{x}))$, i.e.~$F_i(\boldsymbol{x})=\exp(Z_i(\boldsymbol{x}))/\sum_{j\in{\cal Y}} \exp(Z_j(\boldsymbol{x}))$ for $i\in{\cal Y}$.
Finally, we write $C(\boldsymbol{x})$ for the {\bf classification} of the input $\boldsymbol{x}$:
\begin{eqnarray*}
C(\boldsymbol{x}) &=& \argmax_{i\in{\cal Y}} F_i(\boldsymbol{x}),
\end{eqnarray*}
which, since \mbox{softmax} is a monotonic transformation, is equal to $\argmax_{i\in{\cal Y}} Z_i(\boldsymbol{x})$.

An {\bf untargeted adversarial attack} is a (potentially stochastic) mapping $\rho: {\cal X}\to{\cal X}$, aiming to change the output of the classifier, i.e.~$C(\boldsymbol{x}+\rho(\boldsymbol{x}))\neq C(\boldsymbol{x})$, while keeping the {\bf perturbation} $\| \rho(\boldsymbol{x}) \|_p$ small with respect to a particular $\ell_p$ {\bf norm} (most commonly, $p\in\{0,1,2,\infty\}$). Similarly, a {\bf targeted adversarial attack} is a (potentially stochastic) mapping $\psi: {\cal X}\times{\cal Y}\to{\cal X}$, aiming to ensure the classifier outputs a specified class, i.e.~$C(\boldsymbol{x}+\psi(\boldsymbol{x},y))=y$, while keeping the perturbation $\| \psi(\boldsymbol{x},y) \|_p$ small. We call
$\boldsymbol{x}_{\mbox{\tiny adv}}=\boldsymbol{x}+\rho(\boldsymbol{x})$ (or analogously $\boldsymbol{x}_{\mbox{\tiny adv}}=\boldsymbol{x}+\psi(\boldsymbol{x},y)$)  the {\bf adversarial sample} generated by the targeted attack $\rho$ (untargeted attack $\psi$). In practice, we often consider
$\boldsymbol{x}_{\mbox{\tiny adv}}=\mbox{clip}(\boldsymbol{x}+\rho(\boldsymbol{x}), x_{\mbox{\tiny min}}, x_{\mbox{\tiny max}})$ (analogously, $\boldsymbol{x}_{\mbox{\tiny adv}}=\mbox{clip}(\boldsymbol{x}+\psi(\boldsymbol{x},y), x_{\mbox{\tiny min}}, x_{\mbox{\tiny max}})$) to ensure the adversarial samples are in the valid data range.

Finally, the following objects play an important role in the generation of adversarial samples:
\begin{itemize}
\item The {\bf loss function} ${\cal L}: {\cal X}\times{\cal Y} \to {\mathbb R}$ that was used to train the classifier. In many cases, this is the cross-entropy loss: ${\cal L}(\boldsymbol{x},y) = - \log F_y(\boldsymbol{x})$.
\item The {\bf loss gradient}, i.e.~the gradient $\nabla_x {\cal L}(\boldsymbol{x},y)$ of the classifier's loss function with respect to $\boldsymbol{x}$.
\item The {\bf class gradients}, i.e.~either the {\bf logit gradients} $\nabla Z(\boldsymbol{x})$ or the {\bf output gradients} $\nabla F(\boldsymbol{x})$ with respect to $\boldsymbol{x}$.
\end{itemize}

\section{Library Modules} \label{sec:library_modules}

The library is structured as follows:

\dirtree{%
.1 art/.
.2 attacks/.
.3 adversarial\_patch.py.
.3 attack.py.
.3 boundary.py.
.3 carlini.py.
.3 decision\_tree\_attack.py.
.3 deepfool.py.
.3 elastic\_net.py.
.3 fast\_gradient.py.
.3 hclu.py.
.3 hop\_skip\_jump.py.
.3 iterative\_method.py.
.3 newtonfool.py.
.3 projected\_gradient\_descent.py.
.3 saliency\_map.py.
.3 spatial\_transformation.py.
.3 universal\_perturbation.py.
.3 virtual\_adversarial.py.
.3 zoo.py.
.2 classifiers/.
.3 blackbox.py.
.3 catoost.py.
.3 classifier.py.
.3 detector\_classifier.py.
.3 ensemble.py.
.3 GPy.py.
.3 keras.py.
.3 lightgbm.py.
.3 mxnet.py.
.3 pytorch.py.
.3 scikitlearn.py.
.3 tensorflow.py.
.3 xgboost.py.
.2 defences/.
.3 adversarial\_trainer.py.
.3 feature\_squeezing.py.
.3 gaussian\_augmentation.py.
.3 jpeg\_compression.py.
.3 label\_smoothing.py.
.3 pixel\_defend.py.
.3 preprocessor.py.
.3 spatial\_smoothing.py.
.3 thermometer\_encoding.py.
.3 variance\_minimization.py.
.2 detection/.
.3 subsetscanning/.
.4 detector.py.
.4 scanner.py.
.4 scanningops.py.
.4 scoring\_functions.py.
.3 detector.py.
.2 metrics/.
.3 metrics.py.
.3 verification\_decision\_trees.py.
.2 poison\_detection/.
.3 activation\_defence.py.
.3 clustering\_analyzer.py.
.3 ground\_truth\_evaluator.py.
.3 poison\_filtering\_defence.py.
.2 wrappers/.
.3 expectation.py.
.3 query\_efficient\_bb.py.
.3 randomized\_smoothing.py.
.3 wrapper.py
.2 data\_generators.py.
.2 utils.py.
.2 utils\_test.py.
.2 visualization.py.
}

The following sections introduce the general concept, mathematical notation and a formal definition for each module.

\section{Classifiers} \label{sec:classifier}
\bbox{art.classifiers}{}

This module contains the functional API enabling the integration of machine learning models of various libraries into ART. It abstracts from the actual framework in which the classifier is implemented (e.g.\ TensorFlow), and makes the modules for adversarial attacks and defences framework-independent. The following machine learning libraries are supported:

\begin{itemize}
  \item TensorFlow v1: \texttt{art.classifiers.TensorFlowClassifier}
  \item TensorFlow v2: \texttt{art.classifiers.TensorFlowV2Classifier}
  \item Keras: \texttt{art.classifiers.KerasClassifier}
  \item PyTorch \texttt{art.classifiers.PyTorchClassifier}
  \item MXNet: \texttt{art.classifiers.MXClassifier}
  \item Scikit-learn: \texttt{art.classifiers.SklearnClassifier}
  \item XGBoost: \texttt{art.classifiers.XGBoostClassifier}
  \item LightGBM: \texttt{art.classifiers.LightGBMClassifier}
  \item CatBoost: \texttt{art.classifiers.CatBoostARTClassifier}
  \item GPy: \texttt{art.classifiers.GPyGaussianProcessClassifier}
  \item Python function: \texttt{art.classifiers.BlackBoxClassifier}
\end{itemize}

The class \texttt{art.classifiers.EnsembleClassifier} provides support for combining multiple classifiers into an ensemble (Section~\ref{sec:ensemble}).

\subsection{The \texttt{Classifier} Base Class}\label{sec:classifier_base}
\bbox{art.classifiers.Classifier}{art/classifiers/classifier.py}

The \texttt{Classifier} abstract class defines the most basic components and properties of a classifier to attacked with black-box attacks.

This interface grants access to the following properties:

\begin{itemize}

  \item \texttt{clip\_values}: the range of the data as a tuple $(x_{\mbox{\tiny min}}, x_{\mbox{\tiny max}})$.
  \item \texttt{input\_shape}: the shape of one input sample.

\end{itemize}

This interface provides the following methods:

\begin{itemize}

\item {\tt \_\_init\_\_(clip\_values, channel\_index, defences=None, preprocessing=(0, 1))}

Initializes the classifier with the given clip values $(x_{\mbox{\tiny min}},x_{\mbox{\tiny max}})$ and defences.
The preprocessing tuple of the form \texttt{(substractor, divider)} indicate two float values or numpy arrays to be substracted, respectively used to divide the inputs as preprocessing normalisation operations.
These will always be applied by the classifier before performing any operation on data.
The default values \texttt{(0, 1)} correspond to no changes being applied.
The \texttt{defences} parameter is a list of defence instances to be applied in sequence.

\item {\tt predict(x) -> np.ndarray}

Returns predictions of the classifier for the given inputs. If {\tt logits} is {\tt False}, then the class probabilities $F(\cdot)$ are returned, otherwise the class logits $Z(\cdot)$ (predictions before softmax).
The shape of the returned array is $(n,K)$ where $n$ is the number of given samples and $K$ the number of classes.

\item {\tt fit(x, y) -> None}

Fits the classifier to the given data.

\item \texttt{nb\_classes() -> Int}

Returns the number of output classes $K$.

\item {\tt save(filename, path) -> None}

Saves the model to file.

\end{itemize}

\subsubsection{The \texttt{ClassifierNeuralNetwork} Mixin Base Class}\label{sec:ClassifierNeuralNetwork_base}

The interface \texttt{ClassifierNeuralNetwork} is required for neural network classifiers.

The interface grants access to the following properties:

\begin{itemize}

  \item \texttt{channel\_index}: the index of the axis containing the colour channel in the data.
  \item \texttt{layer\_names}: a list with the names of the layers in the model, ordered from input towards output. Note: this list does not include the input and output layers, only the hidden ones. The correctness of this property is not guaranteed and depends on the extent where this information can be extracted from the underlying model.
  \item \texttt{learning\_phase}: the learning phase set by the user for the current classifier (\texttt{True} for training, \texttt{False} for prediction, \texttt{None} if the learning phase has not been set through the library).
  
\end{itemize}

It implements the following methods:

\begin{itemize}

\item {\tt predict(x, batch\_size=128) -> np.ndarray}

Returns predictions of the classifier for the given inputs. If {\tt logits} is {\tt False}, then the class probabilities $F(\cdot)$ are returned, otherwise the class logits $Z(\cdot)$ (predictions before softmax).
The shape of the returned array is $(n,K)$ where $n$ is the number of given samples and $K$ the number of classes.

\item {\tt fit(x, y, batch\_size=128, nb\_epochs=20) -> None}

Fits the classifier to the given data, using the provided batch size and number of epochs.

\item {\tt get\_activations(x, layer) -> np.ndarray}

Computes and returns the values of the activations (outputs) of the specified layer index for the given data \texttt{x}.
The layer index goes from 0 to the total number of internal layers minus one.
The input and output layers are not considered in the total number of available layers.

\item {\tt set\_learning\_phase(train) -> None}

Set the classifier's learning phase to training if \texttt{train} is \texttt{True}, and to prediction otherwise.

\end{itemize}

\subsubsection{The \texttt{ClassifierGradients} Mixin Base Class}\label{sec:ClassifierGradients_base}

The interface \texttt{ClassifierGradients} is required for classifiers to connect with white-box attacks which require the implementation of methods calculating the \texttt{loss\_gradient} and \texttt{class\_gradient}. This base class has to be mixed in with class \texttt{Classifier} and optionally class \texttt{ClassifierNeuralNetwork} to extend the minimum classifier functionality.

\begin{itemize}

\item \texttt{class\_gradient(x, label=None, logits=False) -> np.ndarray}

Returns the gradients of the class probabilities or logits (depending on the value of the {\tt logits} parameter), evaluated at the given inputs.
Specifying a class \texttt{label} (one-hot encoded) will only compute the class gradients for the respective class.
Otherwise, gradients for all classes will be computed.
The shape of the returned array is of the form $(n,K,I)$ when no label is specified, or $(n,1,I)$ for a given label, where $I$ is the shape of the classifier inputs.
If the classifier is equipped with a preprocessor defence, gradients will be computed backwards through it (or approximated in case the defence is not differentiable).

\item \texttt{loss\_gradient(x, y) -> np.ndarray}

Returns the loss gradient evaluated at the given inputs \texttt{x} and \texttt{y}.
The labels are assumed to be one-hot encoded, i.e.\ the label $y$ is encoded as the $y$th standard basis vector $\boldsymbol{e}_y$.
The shape of the returned array is of the form $(n,I)$ under the above notation.
If the classifier is equipped with a preprocessor defence, gradients will be computed backwards through it (or approximated in case the defence is not differentiable).

\end{itemize}

\subsubsection{The \texttt{ClassifierDecisionTree} Mixin Base Class}\label{sec:ClassifierDecisionTree_base}

The interface \texttt{ClassifierDecisionTree} is required for classifiers with decision trees.

\begin{itemize}

\item \texttt{get\_trees() -> [Trees]}

Returns a list of decision trees.

\end{itemize}

\subsection{Ensembles of Classifiers} \label{sec:ensemble}
\bbox{art.classifiers.EnsembleClassifier}{art/classifiers/ensemble.py}

This class aggregates multiple objects of type \texttt{Classifier} as an ensemble.
Its purpose is to allow for attacks to be applied to ensembles. The individual classifiers are expected to be trained when the ensemble is created, and no training procedure is provided through this class.

\section{Classifier Wrappers} \label{sec:classifier_wrappers}
\bbox{art.wrappers}{}

This module contains a functional API which allows user to define wrappers around ART classifiers according to different attack strategies.

\subsection{The \texttt{ClassifierWrapper} Base Class}
\bbox{art.wrappers.ClassifierWrapper}{art/wrappers/wrapper.py}

The \texttt{ClassifierWrapper} class provides a wrapper around \texttt{Classifier} instances. As such, it exposes the same properties and functions as described
in Section \ref{sec:classifier_base}, with exception of the constructor.

\subsection{Expectation over Transformation}\label{sec:wrappers_eot}
\bbox{art.wrappers.ExpectationOverTransformations}{art/wrappers/expectation.py}

The wrapper class \texttt{ExpectationOverTransformations} alters the way how classifier predictions and gradients are computed, namely, it takes the average predictions and gradients, respectively, over a specified number of (random) transformations of the classifier inputs. This technique is a key ingredient to the synthesis of adversarial samples that are robust with respect to (random) transformations of classifier inputs \citep{athalye2017, athalye2018}, occurring either as an intentional defence strategy, or when synthesizing and digitizing adversarial samples in the real world.

\subsection{Query-efficient Black-box Gradient Estimation}\label{sec:wrappers_efficient_bb}
\bbox{art.wrappers.QueryEfficientBBGradientEstimation}{art/wrappers/query\_efficient\_bb.py}

The wrapper class \texttt{QueryEfficientBBGradientEstimation} alters the way how classifier gradients are computed: instead of returning the true gradients, an estimate is computed based on a specified number of classifier predictions. This functionality allows for the emulation of black-box attacks in which the attacker doesn't have direct access to the true classifier gradients, but needs to estimate them based on a certain number of queries to the classifier. 
The \texttt{QueryEfficientBBGradientEstimation} class implements the efficient algorithm proposed in~\citep{ilyas_query-efficient2017}.

\subsection{Randomized Smoothing}\label{sec:wrappers_randomized_smoothing}
\bbox{art.wrappers.RandomizedSmoothing}{art/wrappers/randomized\_smoothing.py}

The wrapper class \texttt{RandomizedSmoothing} alters classifiers that classify well under Gaussian noise to become certifiably robust to adversarial perturbations under the L2 norm \citep{Cohen2019}.

\section{Attacks}\label{sec:attacks}
\bbox{art.attacks}{}

This module contains all the attack methods implemented in ART. They require access to a \texttt{Classifier}, which is the target of the attack.
By using the framework-independent API to access the targeted model, the attack implementation becomes agnostic to the framework used for training the model.

The following attacks are currently implemented in ART:

\begin{itemize}

\item Fast Gradient Sign Method (FGSM)~\citep{goodfellow2014fgsm}, Section~\ref{subsection:FGSM}
\item Basic Iterative Method (BIM)~\citep{kurakin2016a}, Section~\ref{sec:bim}
\item Projected Gradient Descent (PGD)~\citep{madry2017}, Section~\ref{sec:pgd}
\item Jacobian Saliency Map Attack (JSMA)~\citep{papernot2015}, Section~\ref{subsection:JSMA}
\item Carlini \& Wagner attack~\citep{carlini2017}, Section~\ref{subsection:CW}
\item DeepFool~\citep{DBLP:journals/corr/Moosavi-Dezfooli15}, Section~\ref{subsection:deepfool}
\item Universal Perturbation~\citep{Moosavi-Dezfooli16adversarial}, Section~\ref{subsection:universal_perturbation}
\item NewtonFool~\citep{jang2017}, Section~\ref{subsection:newtonfool}
\item Virtual Adversarial Method~\citep{miyato2017virtual}, Section~\ref{subsection:virtual_adversarial}.
\item Spatial Transformation Attack~\citep{engstrom_spatial2017} , Section~\ref{subsection:spatial_transformation}.
\item Elastic Net (EAD) Attack~\citep{chen2017ead}, Section \ref{subsection:ead}.
\item Zeroth-Order-Optimization (ZOO) Attack~\citep{chen2017zoo}, Section~\ref{subsection:zoo}.
\item Boundary Attack~\citep{brendel2017}, Section~\ref{subsection:boundary_attack}.
\item Adversarial Patch~\citep{brown2017patch}, Section~\ref{subsection:adversarial_patch}
\item Decision Tree Attack~\citep{papernot2016transferability}, Section~\ref{subsection:decision_tree_attack}
\item High Confidence Low Uncertainty (HCLU) Attack~\citep{Grosse2018}, Section~\ref{subsection:hclu_attack}
\item HopSkipJump Attack~\citep{Chen2019HopSkipJump}, Section~\ref{subsection:hop_skip_jump_attack}

\end{itemize}

\subsection{The \texttt{Attack} Base Class}
\bbox{art.attacks.Attack}{art/attacks/attack.py}

ART has an abstract class {\tt Attack} in the {\tt art.attacks.attack} module which implements a common interface for any of the particular attacks implemented in the library. The class has an attribute {\tt classifier}, an instance of the {\tt Classifier} class, which is the classifier $C(\boldsymbol{x})$ that the attack aims at. Moreover, the class has the following public methods:
\begin{itemize}
\item {\tt \_\_init\_\_(classifier)}

Initializes the attack with the given classifier.

\item {\tt generate(x, y=None, $\ast\ast$kwargs) -> np.ndarray}

Applies the attack to the given input {\tt x}, using any attack-specific parameters provided in the {\tt $\ast\ast$kwargs} dictionary.
The parameters provided in the dictionary are also set in the attack attributes.
Returns the perturbed inputs in an {\tt np.ndarray} which has the same shape as {\tt x}. The target labels {\tt y} are an {\tt np.ndarray} containing labels for the inputs {\tt x} in one-hot encoding. If the attack is targeted, {\tt y} is required and specifies the target classes. If the attack is untargeted, {\tt y} is optional and overwrites the $C(\boldsymbol{x})$ argument in \eqref{eq:FGSM2}. Note that it is not advisable to provide true labels in the untargeted case, as this may lead to the so-called {\bf label leaking} effect~\citep{kurakin2016a}.

\item {\tt set\_params($\ast\ast$kwargs) -> bool}

Initializes attack-specific hyper-parameters provided in the {\tt $\ast\ast$kwargs} dictionary; returns ``True'' if the hyper-parameters were valid and the initialization successful, ``False'' otherwise.
\end{itemize}

\subsection{FGSM} \label{subsection:FGSM}
\bbox{art.attacks.FastGradientMethod}{art/attacks/fast\_gradient.py}

The Fast Gradient Sign Method (FGSM)~\citep{goodfellow2014fgsm} works both in targeted and untargeted settings, and aims at controlling either the $\ell_1$, $\ell_2$ or $\ell_\infty$ norm of the adversarial perturbation.
In the targeted case and for the $\ell_\infty$ norm, the adversarial perturbation generated by the FGSM attack is given by
\begin{eqnarray*}
\psi(\boldsymbol{x}, y) = - \epsilon \cdot \mbox{sign}(\nabla_{\boldsymbol{x}} {\cal L}(\boldsymbol{x},y)) \label{eq:FGSM1}
\end{eqnarray*}
where $\epsilon>0$ is the {\bf attack strength} and $y$ is the target class specified by the attacker.
The adversarial sample is given by
\begin{eqnarray*}
\boldsymbol{x}_{\mbox{\tiny adv}} =  \mbox{clip}(\boldsymbol{x}+\psi(\boldsymbol{x}, y), x_{\mbox{\tiny min}}, x_{\mbox{\tiny max}}).
\end{eqnarray*}

Intuitively, the attack transforms the input $\boldsymbol{x}$ to reduce the classifier's loss when classifying it as $y$.
For the $\ell_p$ norms with $p=1,2$, the adversarial perturbation is calculated as
\begin{eqnarray*}
\psi(\boldsymbol{x}, y) &=& \epsilon \cdot \frac{\nabla_{\boldsymbol{x}} {\cal L}(\boldsymbol{x},y)}{\|\nabla_{\boldsymbol{x}} {\cal L}(\boldsymbol{x},y)\|_p}.
\end{eqnarray*}

Note that ``Sign'' in the attack name ``FGSM'' refers to the specific calculation for the $\ell_\infty$ norm (for which the attack was originally introduced).
Sometimes the modified attacks for the $\ell_1$ and $\ell_2$ norms are referred to as ``Fast Gradient Methods'' (FGM), nevertheless we will also refer to them as FGSM for simplicity.
The untargeted version of the FGSM attack is devised as
\begin{eqnarray}
\rho(\boldsymbol{x}) = -\psi(\boldsymbol{x}, C(\boldsymbol{x})), \label{eq:FGSM2}
\end{eqnarray}
i.e.~the input $\boldsymbol{x}$ is transformed to increase the classifier's loss when continuing to classify it as $C(\boldsymbol{x})$.

ART also implements an extension of the FGSM attack, in which the {\bf minimum perturbation} is determined for which $C(\boldsymbol{x}_{\mbox{\tiny adv}}) \neq C(\boldsymbol{x})$.
This modification takes as input two extra floating-point parameters: $\epsilon_{\mbox{\tiny step}}$ and $\epsilon_{\mbox{\tiny max}}$.
It sequentially performs the standard FGSM attack with strength $\epsilon=k\cdot\epsilon_{\mbox{\tiny step}}$ for $k=1,2,\ldots$ until either the attack is successful (and the resulting adversarial sample $\boldsymbol{x}_{\mbox{\tiny adv}}$ is returned), or $k\cdot\epsilon_{\mbox{\tiny step}} > \epsilon_{\mbox{\tiny max}}$, in which case the attack has failed.

The main advantage of FGSM is that it is very efficient to compute: only one gradient evaluation is required, and the attack can be applied straight-forward to a batch of inputs.
This makes the FGSM (or variants thereof) a popular choice for adversarial training (see Section~\ref{subsec:Adversarial_training}) in which a large number of adversarial samples needs to be generated.

The strength of the FGSM attack depends on the choice of the parameter $\epsilon$.
If $\epsilon$ is too small, then $C(\boldsymbol{x}_{\mbox{\tiny adv}})$ might not differ from $C(\boldsymbol{x})$.
On the other hand, the norm $\|\boldsymbol{x} - \boldsymbol{x}_{\mbox{\tiny adv}} \|_p$ grows linearly with $\epsilon$.
When choosing $\epsilon$, it is particularly important to consider the actual data range $[x_{\mbox{\tiny min}}, x_{\mbox{\tiny max}}]$.

\subsection{Basic Iterative Method} \label{sec:bim}
\bbox{art.attacks.BasicIterativeMethod}{art/attacks/iterative\_method.py}

The Basic Iterative Method (BIM)~\citep{kurakin2016a} is a straightforward extension of FGSM that applies the attack multiple times, iteratively.
In its original version, this attack differs from FGSM in that it is targeted towards the least likely class for a given sample, i.e.\ the class for which the model outputs the lowest score.
The library implementation uses the same options for targets as for FGM (least likely class is not the default, but can be specified by the user).
Like in the case of FGM, BIM is limited by a total attack budget $\epsilon$, and it has an additional parameter $\epsilon_{\mbox{{\small step}}}$ determining the step size at each iteration.
At each step, the result of the attack is projected back onto the $\epsilon$-size ball centered around the original input.
The attack is implemented as a special case of Projected Gradient Descent with $L_{\infty}$ norm perturbation and no random initialisation.

\subsection{Projected Gradient Descent} \label{sec:pgd}
\bbox{art.attacks.ProjectedGradientDescent}{art/attacks/projected\_gradient\_descent.py}

Projected Gradient Descent (PGD)~\citep{madry2017} is also an iterative extension of FGSM and very similar to BIM.
The main difference with BIM resides in the fact that PGD projects the attack result back on the $\epsilon$-norm ball around the original input at each iteration of the attack.

\subsection{Jacobian Saliency Map Attack (JSMA)} \label{subsection:JSMA}
\bbox{art.attacks.SaliencyMapMethod}{art/attacks/saliency\_map.py}

The Jacobian-based Saliency Map Attack (JSMA)~\citep{papernot2015} is a targeted attack which aims at controlling the $\ell_0$ norm, i.e.~the number of components of $\boldsymbol{x}$ that are being modified when crafting the adversarial sample $\boldsymbol{x}_{\mbox{\tiny adv}}$.
The attack iteratively modifies individual components of $\boldsymbol{x}$ until either the targeted misclassification is achieved or the total number of modified components exceeds a specified budget.

Details are outlined in Algorithm~\ref{alg:JSMA}. To simplify notation, we let $N$ denote the total number of components of $\boldsymbol{x}$ and refer to individual components using subscripts: $x_i$ for $i=1,\ldots,N$.
The key step is the computation of the saliency map (line \ref{alg:JSMA:compute_saliency_map}), which is outlined in Algorithm~\ref{alg:JSMA:saliency_map}.
Essentially, the saliency map determines the components $i_{\mbox{\tiny max}}$, $j_{\mbox{\tiny max}}$ of $\boldsymbol{x}$ to be modified next based on how much this would increase the probability $F_y(\boldsymbol{x})$ of the target class $y$ (captured in the sum of partial derivatives $\alpha$), and decrease the sum of probabilities over all other classes (captured in the sum of partial derivatives $\beta$). In particular, line~\ref{alg:JSMA:saliency_map_criterion}  in Algorithm~\ref{alg:JSMA:saliency_map} ensures that $\alpha$ is positive, $\beta$ is negative and $|\alpha\cdot\beta|$ is maximal over all pairs in the search space. Considering pairs instead of individual components is motivated by the observation that this increases the chances of satisfying the conditions $\alpha>0$ and $\beta<0$  (\citep{papernot2015}, page 9). In our implementation of the saliency map, we exploit that $\beta$ can be expressed as
\begin{eqnarray*}
\beta &=& - \left( \frac{\partial F_y(\boldsymbol{x})}{\partial x_i} \, + \,  \frac{\partial F_y(\boldsymbol{x})}{\partial x_j} \right)
\end{eqnarray*}
(due to $\sum_{k\in{\cal Y}\setminus\{y\}} F_k(\boldsymbol{x})=1-F_y(\boldsymbol{x})$), and thus $i$, $j$ are the two largest components of $\nabla F_y(\boldsymbol{x})$.

Line~\ref{alg:JSMA:perturb_ij} in Algorithm~\ref{alg:JSMA} applies the perturbations to the components determined by the saliency map.
The set $\Omega_{\mbox{\tiny search}}$ keeps track of all the components that still can be modified (i.e.~the clipping value $x_{\mbox{\tiny max}}$ has not been exceeded), and $\Omega_{\mbox{\tiny modified}}$ of all the components that have been modified at least once.
The algorithm terminates if the attack has succeeded (i.e.~$C(\boldsymbol{x})= y$), the number of modified components has exhausted the budget (i.e.~$|\Omega_{\mbox{\tiny modified}}| > \lfloor \gamma \cdot N\rfloor$), less than $2$ components can still be modified (i.e.~$|\Omega_{\mbox{\tiny search}}|<2$), or the saliency map returns the saliency score $s_{\mbox{\tiny max}}=-\infty$ (indicating it hasn't succeeded in determining components satisfying the conditions in line~\ref{alg:JSMA:saliency_map_criterion} of Algorithm~\ref{alg:JSMA:saliency_map}).

\begin{algorithm}[h]
  \caption{JSMA method}
   \label{alg:JSMA}
\begin{algorithmic}[1]
\REQUIRE $\mbox{ }$ \\
$\boldsymbol{x}$: Input to be adversarially perturbed \\
$y$: Target label  \\
$\theta$: Amount of perturbation per step and feature (assumed to be positive; see discussion in text) \\
$\gamma$: Maximum fraction of features to be perturbed (between $0$ and $1$)
\STATE $\Omega_{\mbox{\tiny modified}} \leftarrow \emptyset$
\STATE $\Omega_{\mbox{\tiny search}} \leftarrow \{ i \in\{1,\ldots,N\}: \ x_i \neq x_{\mbox{\tiny max}} \}$ \label{alg:JSMA:search_space_init}
\WHILE{$C(\boldsymbol{x})\neq y$ and $|\Omega_{\mbox{\tiny search}}|>1$}
\STATE{$(s_{\mbox{\tiny max}}, i, j) \leftarrow \mbox{\tt saliency\_map}(\boldsymbol{x}, y, \Omega_{\mbox{\tiny search}})$} \label{alg:JSMA:compute_saliency_map}
\STATE  $\Omega_{\mbox{\tiny modified}} \leftarrow \Omega_{\mbox{\tiny modified}} \cup \{i,j\}$
\IF{$s_{\mbox{\tiny max}}=-\infty$ or $|\Omega_{\mbox{\tiny modified}}| > \lfloor \gamma \cdot N\rfloor$} \STATE break \ENDIF
\STATE  $x_i \leftarrow \mbox{clip}(x_i + \theta,  x_{\mbox{\tiny min}},  x_{\mbox{\tiny max}}) $,
$x_j \leftarrow \mbox{clip}(x_j + \theta,  x_{\mbox{\tiny min}},  x_{\mbox{\tiny max}}) $ \label{alg:JSMA:perturb_ij}
\STATE  $\Omega_{\mbox{\tiny search}} \leftarrow \Omega_{\mbox{\tiny search}} \setminus \{ k \in\{i,j\}: \ x_k = x_{\mbox{\tiny max}} \}$ \label{alg:JSMA_search_space_update}
\ENDWHILE
\STATE  $\boldsymbol{x}_{\mbox{\tiny adv}} \leftarrow \boldsymbol{x}$
\ENSURE  $\mbox{ }$ \\
Adversarial sample $\boldsymbol{x}_{\mbox{\tiny adv}}$.
\end{algorithmic}
\end{algorithm}

\begin{algorithm}[h]
  \caption{JSMA \mbox{\tt saliency\_map}}
   \label{alg:JSMA:saliency_map}
\begin{algorithmic}[1]
\REQUIRE $\mbox{ }$ \\
$\boldsymbol{x}$: Input to be adversarially perturbed \\
$y$: Target label \\
$\Omega_{\mbox{\tiny search}}$: Set of indices of $\boldsymbol{x}$ to be explored.
\STATE $s_{\mbox{\tiny max}} \leftarrow -\infty$, $i_{\mbox{\tiny max}} \leftarrow \mbox{\tt None}$, $j_{\mbox{\tiny max}} \leftarrow \mbox{\tt None}$
\FOR{ each subset $\{i,j\}\subset\Omega_{\mbox{\tiny search}}$ with $i<j$}
\STATE { Compute \begin{eqnarray*}
\alpha \, =\, \frac{\partial F_y(\boldsymbol{x})}{\partial x_i} \, + \,  \frac{\partial F_y(\boldsymbol{x})}{\partial x_j}\,, \
\beta \, = \, \sum_{k\in{\cal Y}\setminus\{y\}} \left( \frac{\partial F_k(\boldsymbol{x})}{\partial x_i} \, + \,  \frac{\partial F_k(\boldsymbol{x})}{\partial x_j} \right).
\end{eqnarray*}}
\IF{$\alpha>0$ and $\beta<0$ and $|\alpha\cdot\beta|>s_{\mbox{\tiny max}}$} \label{alg:JSMA:saliency_map_criterion}
\STATE $s_{\mbox{\tiny max}} \leftarrow |\alpha\cdot\beta|$, $i_{\mbox{\tiny max}} \leftarrow i$, $j_{\mbox{\tiny max}} \leftarrow j$
\ENDIF
\ENDFOR
\ENSURE  $\mbox{ }$ \\
Maximum saliency score and selected components $(s_{\mbox{\tiny max}}, i_{\mbox{\tiny max}}, j_{\mbox{\tiny max}})$.
\end{algorithmic}
\end{algorithm}

\begin{itemize}
\item Algorithms~\ref{alg:JSMA} and~\ref{alg:JSMA:saliency_map} describe the method for positive values of the input parameter $\theta$, resulting in perturbations where the components of $\boldsymbol{x}$ are increased while crafting $\boldsymbol{x}_{\mbox{\tiny adv}}$.
In principle, it is possible to also use negative values for $\theta$.
In fact, the experiments in~\citep{papernot2015} (Section IV D) suggest that using a negative $\theta$ results in perturbations that are harder to detect by the human eye\footnote[5]{Which however might be specific to the MNIST data set that was used in those experiments.}.
The implementation in ART supports both positive and negative $\theta$, however note that for negative $\theta$ the following changes apply:
\begin{itemize}
\item In Algorithm~\ref{alg:JSMA}, line~\ref{alg:JSMA:search_space_init}: change the condition $x_i \neq x_{\mbox{\tiny max}}$ to $x_i \neq x_{\mbox{\tiny min}}$.
\item In Algorithm~\ref{alg:JSMA}, line~\ref{alg:JSMA_search_space_update}: change the condition $x_k \neq x_{\mbox{\tiny max}}$ to $x_k \neq x_{\mbox{\tiny min}}$.
\item In Algorithm~\ref{alg:JSMA:saliency_map}, line~\ref{alg:JSMA:saliency_map_criterion}: change the conditions $\alpha>0$ and $\beta<0$ to $\alpha<0$ and $\beta>0$, respectively.
\end{itemize}
\item While Algorithm~\ref{alg:JSMA:saliency_map} outlines the computation of the saliency map based on the partial derivates of the classifier probabilities $F(\boldsymbol{x})$, one can, in principle, consider the partial derivates of the classifier logits $Z(\boldsymbol{x})$ instead.
As discussed in~\citep{carlini2017}, it appears that both variants have been implemented and used in practice.
The current implementation in ART uses the partial derivates of the classifier probabilities which, as explained above, can be performed particularly efficiently.
\end{itemize}

\subsection{Carlini \& Wagner $\ell_2$ attack}\label{subsection:CW}
\bbox{art.attacks.CarliniL2Method}{art/attacks/carlini.py}

The Carlini \& Wagner (C\&W) $\ell_2$ attack~\citep{carlini2017} is a targeted attack which aims to minimize the $\ell_2$ norm of adversarial perturbations. Below we will discuss how to perform untargeted attacks. For a fixed input $\boldsymbol{x}$, a target label $y$ and a {\bf confidence} parameter $\kappa\geq 0$, consider the objective function
\begin{eqnarray}
L(c, \boldsymbol{x}') &:=& \|\boldsymbol{x}'-\boldsymbol{x}\|_2^2 + c \cdot \ell(\boldsymbol{x}') \label{eq:CW:objective_function}
\end{eqnarray}
where
\begin{eqnarray}
\ell(\boldsymbol{x}') &=& \max\Big( \max \big\{Z_i(\boldsymbol{x}'): \ i\in {\cal Y}\setminus\{ y \} \big\} - Z_y(\boldsymbol{x}') + \kappa, \, 0 \Big).
 \label{eq:CW:loss_function}
\end{eqnarray}
Note that $\ell(\boldsymbol{x}')=0$ if and only if $C(\boldsymbol{x}')=y$ and the logit $Z_y(\boldsymbol{x}')$ exceeds any other logit $Z_i(\boldsymbol{x}')$ by at least $\kappa$, thus relating to the classifier's confidence in the output $C(\boldsymbol{x}')$.
The C\&W attack aims at finding the smallest $c$ for which the $\boldsymbol{x}'$ minimizing $L(\boldsymbol{x}',c)$ is such that $\ell(\boldsymbol{x}')=0$.
This can be regarded as the optimal trade-off between achieving the adversarial target while keeping the adversarial perturbation $\|\boldsymbol{x}'-\boldsymbol{x}\|_2^2$ as small as possible. In the ART implementation, binary search is used for finding such $c$.

Details are outlined in Algorithm \ref{alg:CW}. Note that, in lines \ref{alg:CW:input_preprocessing_1}-\ref{alg:CW:input_preprocessing_2}, the components of $\boldsymbol{x}$ are mapped from $[x_{\mbox{\tiny min}}, x_{\mbox{\tiny max}}]$ onto ${\mathbb R}$, which is the space in which the adversarial sample is created. Working in ${\mathbb R}$ avoids the need for clipping $\boldsymbol{x}$ during the process. The output $\boldsymbol{x}_{\mbox{\tiny adv}}$ is transformed back to the $[x_{\mbox{\tiny min}}, x_{\mbox{\tiny max}}]$ range in lines \ref{alg:CW:output_postprocessing_1}-\ref{alg:CW:output_postprocessing_2}.
Algorithm \ref{alg:CW} relies on two helper functions:
\begin{itemize}
\item {\tt minimize\_objective} (line \ref{alg:CW:minimize_objective}): this function aims at minimizing the objective function (\ref{eq:CW:objective_function}) for the given value of $c$. In the ART implementation, the minimization is performed using gradient descent with binary line search.\footnote[1]{The original implementation in \citep{carlini2017} used the Adam optimizer for the minimization problem and implemented an optional early stopping criterion.} The maximum number of halvings and doublings of the search step is specified via the parameters {\tt max\_halving} and {\tt max\_doubling}, respectively. The search is aborted if the number of iterations exceeds the value given by the {\tt max\_iter} parameter.
During the minimization, among all samples $\boldsymbol{x}'$ for which $\ell(\boldsymbol{x}')=0$, the one with the smallest norm $\|\boldsymbol{x}'-\boldsymbol{x}\|_2^2$ is retained and returned at the end of the minimization.
\item {\tt update} (line \ref{alg:CW:c_updates}): this function updates the parameters $c_{\mbox{\tiny lower}}$, $c$, $c_{\mbox{\tiny double}}$ used in the binary search. Specifically, if an adversarial sample with $\ell(\boldsymbol{x}_{\mbox{\tiny new}})=0$ was found for the previous value of $c$, then $c_{\mbox{\tiny lower}}$ remains unchanged, $c$ is set to $(c+c_{\mbox{\tiny lower}})/2$ and $c_{\mbox{\tiny double}}$ is set to {\tt False}. If  $\ell(\boldsymbol{x}_{\mbox{\tiny new}})\neq 0$, then $c_{\mbox{\tiny lower}}$ is set to $c$, and $c$ is either set to $2\cdot c$ (if $c_{\mbox{\tiny double}}$ is {\tt True}) or to  $c+(c-c_{\mbox{\tiny lower}})/2$
(if $c_{\mbox{\tiny double}}$ is {\tt False}).\footnote[2]{This is a slight simplification of the original implementation in \citep{carlini2017}.} The binary search is abandoned if $c$ exceeds the upper bound $c_{\mbox{\tiny upper}}$ (line \ref{alg:CW:binary_search_while}).
\end{itemize}

The {\bf untargeted} version of the C\&W attack aims at changing the original classification, i.e.~constructing an adversarial sample $\boldsymbol{x}_{\mbox{\tiny adv}}$ with the only constraint that $C(\boldsymbol{x}_{\mbox{\tiny adv}})\neq C(\boldsymbol{x})$. The only difference in the algorithm is that the objective function
(\ref{eq:CW:loss_function}) is modified as follows:
\begin{eqnarray}
\ell(\boldsymbol{x}') &=& \max\Big(Z_y(\boldsymbol{x}') - \max \big\{Z_i(\boldsymbol{x}'): \  i\in {\cal Y}\setminus\{ y \} \big\} + \kappa, \, 0 \Big)
 \label{eq:CW:loss_function+untargeted}
\end{eqnarray}
where $y$ is the original label of $\boldsymbol{x}$ (or the prediction $C(\boldsymbol{x})$ thereof). Thus, $\ell(\boldsymbol{x}')=0$ if and only if there exists a label $i\in{\cal Y}\setminus\{y\}$ such that the logit $Z_i(\boldsymbol{x})$ exceeds the logit $Z_y(\boldsymbol{x})$ by at least $\kappa$.

\begin{algorithm}
  \caption{Carlini \& Wagner's $\ell_2$ attack (targeted)}
   \label{alg:CW}
\begin{algorithmic}[1]
\REQUIRE $\mbox{ }$ \\
$\boldsymbol{x}$: Input to be adversarially perturbed \\
$y$: Target label \\
$\gamma$: constant to avoid over-/underflow in (arc)tanh computations \\
$c_{\mbox{\tiny init}}$: initialization of the binary search constant \\
$c_{\mbox{\tiny upper}}$: upper bound for the binary search constant \\
$b_{\mbox{\tiny steps}}$: number of binary search steps to be performed
\STATE $\boldsymbol{x} \leftarrow (\boldsymbol{x}-x_{\mbox{\tiny min}}) / (x_{\mbox{\tiny max}}-x_{\mbox{\tiny min}})$ \label{alg:CW:input_preprocessing_1}
\STATE $\boldsymbol{x} \leftarrow \mbox{arctanh}( ((2\cdot \boldsymbol{x}) - 1) \cdot \gamma)$ \label{alg:CW:input_preprocessing_2}
\STATE $\boldsymbol{x}_{\mbox{\tiny adv}} \leftarrow \boldsymbol{x}$
\STATE $c_{\mbox{\tiny lower}}\leftarrow 0$, $c \leftarrow c_{\mbox{\tiny init}}$,  $c_{\mbox{\tiny double}} \leftarrow \mbox{\tt True}$
\STATE $l_{\mbox{\tiny min}} \leftarrow \infty$
\WHILE{ $b_{\mbox{\tiny steps}}>0$ and $c<c_{\mbox{\tiny upper}}$} \label{alg:CW:binary_search_while}
\STATE $\boldsymbol{x}_{\mbox{\tiny new}} \leftarrow \mbox{\tt minimize\_objective}(c)$ \label{alg:CW:minimize_objective}
\IF{$\ell(\boldsymbol{x}_{\mbox{\tiny new}})=0$ and  $\|\boldsymbol{x}_{\mbox{\tiny new}}-\boldsymbol{x}\|_2^2 < l_{\mbox{\tiny min}}$}
\STATE $l_{\mbox{\tiny min}} \leftarrow \|\boldsymbol{x}_{\mbox{\tiny new}}-\boldsymbol{x}\|_2^2$
\STATE $\boldsymbol{x}_{\mbox{\tiny adv}} \leftarrow  \boldsymbol{x}_{\mbox{\tiny new}}$
\ENDIF
\STATE $(c_{\mbox{\tiny lower}}, c, c_{\mbox{\tiny double}}) \leftarrow \mbox{\tt update}(c, c_{\mbox{\tiny double}}, \ell(\boldsymbol{x}_{\mbox{\tiny new}}))$ \label{alg:CW:c_updates}
\STATE  $b_{\mbox{\tiny steps}} \leftarrow b_{\mbox{\tiny steps}}-1$
\ENDWHILE
\STATE $\boldsymbol{x}_{\mbox{\tiny adv}} \leftarrow (\mbox{tanh}(\boldsymbol{x}_{\mbox{\tiny adv}})/\gamma+1)/2$ \label{alg:CW:output_postprocessing_1}
\STATE $\boldsymbol{x}_{\mbox{\tiny adv}} \leftarrow \boldsymbol{x}_{\mbox{\tiny adv}}\cdot(x_{\mbox{\tiny max}}-x_{\mbox{\tiny min}})+x_{\mbox{\tiny min}}$ \label{alg:CW:output_postprocessing_2}
\ENSURE  $\mbox{ }$ \\
Adversarial sample $\boldsymbol{x}_{\mbox{\tiny adv}}$.
\end{algorithmic}
\end{algorithm}

\subsection{Carlini \& Wagner $\ell_{\infty}$ attack}\label{subsection:CWLInf}
\bbox{art.attacks.CarliniLInfMethod}{art/attacks/carlini.py}

The Carlini \& Wagner (C\&W) LInf attack aims at finding an adversarial sample $\boldsymbol{x}'$ satisfying the objective $\ell(\boldsymbol{x}') = 0$ (with $\ell$ same as for the $\ell_2$ attack, confer (\ref{eq:CW:loss_function})) while obeying the constraint $\| \boldsymbol{x} - \boldsymbol{x}' \|_\infty \leq \epsilon$ for a given $\epsilon>0$.
The trick for solving this problem is to define an invertible transformation
\begin{eqnarray*}
\boldsymbol{x} \mapsto t(\boldsymbol{x})
\end{eqnarray*}
which element-wise maps the components of $\boldsymbol{x}$ from $[x_{\mbox{\tiny min}},  x_{\mbox{\tiny max}}]$ onto $\mathbb{R}$ and ensures that
for arbitrary $\boldsymbol{z}$ in the domain of $t^{-1}$
\begin{eqnarray*}
\| \boldsymbol{x} - t^{-1}(\boldsymbol{z}) \|_\infty &\leq& \epsilon.
\end{eqnarray*}
This is accomplished by defining
\begin{eqnarray*}
t(\boldsymbol{x}) &:=& \mbox{arctanh}( ((2\cdot \tilde{\boldsymbol{x}}) - 1) \cdot \gamma)
\end{eqnarray*}
where the constant $\gamma=0.999999$ prevents numerical overflow and
\begin{eqnarray*}
\tilde{\boldsymbol{x}} &=&
\frac{\boldsymbol{x} - \mbox{max}(\boldsymbol{x}-\epsilon, x_{\mbox{\tiny min}})}
{  \mbox{min}(\boldsymbol{x}+\epsilon, x_{\mbox{\tiny max}})  -  \mbox{max}(\boldsymbol{x}-\epsilon, x_{\mbox{\tiny min}})}.
\end{eqnarray*}
The C\&W LInf attack then simply solves the unconstrained optimization problem
\begin{eqnarray*}
\boldsymbol{x}_{\mbox{\tiny adv}} &=& t^{-1} \big( \argmin_{\boldsymbol{z}} \ell(t^{-1}(\boldsymbol{z})) \big).
\end{eqnarray*}
Note: this version of the LInf attack differs from the one introduced in \citep{carlini2017} which is computationally more involved and doesn't ensure that
$\| \boldsymbol{x} - \boldsymbol{x}_{\mbox{\tiny adv}} \|_\infty \leq \epsilon$ .
For solving the minimization problem, the same binary line search approach as in the L2 attack is used. The definition of an untargeted attack is analogous to the L2 case.

\paragraph{Implementation details}
The C\&W LInf attack is implemented in the {\tt CarliniLInfMethod} class, with the following attributes:
\begin{itemize}
\item {\tt confidence}: $\kappa$ value to be used in Objective (\ref{eq:CW:loss_function}).
\item {\tt targeted}: Specifies whether the attack should be targeted ({\tt True}) or not ({\tt False}).
\item {\tt learning\_rate}: The initial step size to be used for binary line search in \mbox{\tt minimize\_objective} (line \ref{alg:CW:minimize_objective}).
\item {\tt max\_iter}: Maximum number of iterations to be used for binary line search in \mbox{\tt minimize\_objective}.
\item {\tt max\_halving}: Maximum number of step size halvings per iteration in \mbox{\tt minimize\_objective}.
\item {\tt max\_halving}: Maximum number of step size doublings per iteration in \mbox{\tt minimize\_objective}.
\item {\tt eps}: The $\epsilon$ value in the $\ell_\infty$ norm constraint.
\item {\tt batch\_size}: Internal size of the batches on which adversarial samples are generated.
\end{itemize}

\subsection{DeepFool} \label{subsection:deepfool}
\bbox{art.attacks.DeepFool}{art/attacks/deepfool.py}

DeepFool~\citep{DBLP:journals/corr/Moosavi-Dezfooli15} is an untargeted attack which aims, for a given input $\boldsymbol{x}$, to find the nearest decision boundary in $\ell_2$ norm\footnote[3]{While \citep{DBLP:journals/corr/Moosavi-Dezfooli15} also explains how to adapt the algorithm to minimize the $\ell_1$ or $\ell_\infty$ norms of adversarial perturbations by modifying the expressions in (\ref{alg:DeepFool:eq1})-(\ref{alg:DeepFool:eq2}), this is currently not implemented in ART.}. Implementation details are provided in Algorithm~\ref{alg:DeepFool}.
The basic idea is to project the input onto the nearest decision boundary; since the decision boundaries are non-linear, this is done iteratively. DeepFool often results in adversarial samples that lie exactly on a decision boundary; in order to push the samples over the boundaries and thus change their classification, the final adversarial perturbation $\boldsymbol{x}_{\mbox{\tiny adv}}-\boldsymbol{x}$ is multiplied by a factor $1+\epsilon$ (see line~\ref{alg:DeepFool_overshoot}).

\begin{algorithm}
  \caption{DeepFool attack}
   \label{alg:DeepFool}
\begin{algorithmic}[1]
\REQUIRE $\mbox{ }$ \\
$\boldsymbol{x}$: Input to be adversarially perturbed \\
$i_{\mbox{\tiny max}}$: Maximum number of projection steps \\
$\epsilon$: Overshoot parameter (must be $\geq 0$)
\STATE $\boldsymbol{x}_{\mbox{\tiny adv}} \leftarrow \boldsymbol{x}$
\STATE $i \leftarrow 0$
\WHILE{ $i < i_{\mbox{\tiny max}}$ and $C(\boldsymbol{x})=C(\boldsymbol{x}_{\mbox{\tiny adv}})$} \label{alg:DeepFool:while}
\STATE Compute
 \begin{eqnarray}
l &\leftarrow& \argmin_{k\in{\cal Y}\setminus\{C(\boldsymbol{x}) \}} \frac{\big|Z_k(\boldsymbol{x}_{\mbox{\tiny adv}}) - Z_{C(\boldsymbol{x})}(\boldsymbol{x}_{\mbox{\tiny adv}})\big|}{\big\| \nabla Z_k(\boldsymbol{x}_{\mbox{\tiny adv}}) - \nabla Z_{C(\boldsymbol{x})}(\boldsymbol{x}_{\mbox{\tiny adv}})\big\|_2}, \label{alg:DeepFool:eq1}\\
\boldsymbol{x}_{\mbox{\tiny adv}} &\leftarrow& \boldsymbol{x}_{\mbox{\tiny adv}} +
\frac{\big|Z_l(\boldsymbol{x}_{\mbox{\tiny adv}}) - Z_{C(\boldsymbol{x})}(\boldsymbol{x}_{\mbox{\tiny adv}})\big|}{\big\| \nabla Z_l(\boldsymbol{x}_{\mbox{\tiny adv}}) - \nabla Z_{C(\boldsymbol{x})}(\boldsymbol{x}_{\mbox{\tiny adv}})\big\|_2^2}\cdot \big( \nabla Z_l(\boldsymbol{x}_{\mbox{\tiny adv}}) - \nabla Z_{C(\boldsymbol{x})}(\boldsymbol{x}_{\mbox{\tiny adv}})\big).  \label{alg:DeepFool:eq2}
\end{eqnarray}
\STATE  $\boldsymbol{x}_{\mbox{\tiny adv}} \leftarrow \mbox{clip}(\boldsymbol{x}_{\mbox{\tiny adv}},  x_{\mbox{\tiny min}},  x_{\mbox{\tiny max}})$
\STATE $i \leftarrow i+1$
\ENDWHILE
\STATE  $\boldsymbol{x}_{\mbox{\tiny adv}} \leftarrow \mbox{clip}(\boldsymbol{x} + (1+\epsilon)\cdot(\boldsymbol{x}_{\mbox{\tiny adv}}-\boldsymbol{x}),  x_{\mbox{\tiny min}},  x_{\mbox{\tiny max}})$ \label{alg:DeepFool_overshoot}
\ENSURE  $\mbox{ }$ \\
Adversarial sample $\boldsymbol{x}_{\mbox{\tiny adv}}$.
\end{algorithmic}
\end{algorithm}

\subsection{Universal Adversarial Perturbations} \label{subsection:universal_perturbation}
\bbox{art.attacks.UniversalPerturbation}{art/attacks/universal\_perturbation.py}

Universal adversarial perturbations~\citep{Moosavi-Dezfooli16adversarial} are a special type of untargeted attacks, aiming to create a constant perturbation $\boldsymbol{\rho}$ that successfuly alters the classification of a specified fraction of inputs.
The universal perturbation is crafted using a given untargeted attack $\rho(\cdot)$.
Essentially, as long as the target fooling rate has not been achieved or the maximum number of iterations has not been reached, the algorithm iteratively adjusts the universal perturbation by adding refinements which help to perturb additional samples from the input set; after each iteration, the universal perturbation is projected into the $\ell_p$ ball with radius $\epsilon$ in order to control the attack strength.
Details are provided in Algorithm~\ref{alg:Universal}.

\begin{algorithm}
  \caption{Universal Adversarial Perturbation}
   \label{alg:Universal}
\begin{algorithmic}[1]
\REQUIRE $\mbox{ }$ \\
$\boldsymbol{X}$: Set of inputs to be used for constructing the universal adversarial perturbation \\
$\rho(\cdot)$: Adversarial attack to be used \\
$\delta$: Attack failure tolerance ($1-\delta$ is the target fooling rate) \\
$\epsilon$: Attack step size \\
$p$: Norm of the adversarial perturbation \\
$i_{\mbox{\tiny max}}$: Maximum number of iterations \\
\STATE $\boldsymbol{\rho} \leftarrow \boldsymbol{0}$
\STATE $r_{\mbox{\tiny fool}} \leftarrow 0$
\STATE $i \leftarrow 0$
\WHILE{$r_{\mbox{\tiny fool}}<1-\delta$ and $i < i_{\mbox{\tiny max}}$}
\FOR{$\boldsymbol{x}\in\boldsymbol{X}$ in random order}
\IF{$C(\boldsymbol{x}+\boldsymbol{\rho})=C(\boldsymbol{x})$}
\STATE  $\boldsymbol{x}_{\mbox{\tiny adv}}\leftarrow \boldsymbol{x} + \rho(\boldsymbol{x}+\boldsymbol{\rho})$
\IF{$C(\boldsymbol{x}_{\mbox{\tiny adv}})\neq C(\boldsymbol{x})$}
\STATE $\boldsymbol{\rho} \leftarrow \mbox{project}(\boldsymbol{x}_{\mbox{\tiny adv}} - \boldsymbol{x} + \boldsymbol{\rho}, p, \epsilon)$  \label{alg:Universal:project}
\ENDIF
\ENDIF
\ENDFOR
\STATE $r_{\mbox{\tiny fool}} \leftarrow \frac{1}{|\boldsymbol{X}|}\sum_{\boldsymbol{x}\in\boldsymbol{X}}{\mathbb I}(C(\boldsymbol{x})\neq C(\boldsymbol{x}+\boldsymbol{\rho})) $
\STATE $i \leftarrow i+1$
\ENDWHILE
\ENSURE  $\mbox{ }$ \\
Adversarial samples $\boldsymbol{x}_{\mbox{\tiny adv}}=\boldsymbol{x}+\boldsymbol{\rho}$ for $\boldsymbol{x}\in\boldsymbol{X}$.
\end{algorithmic}
\end{algorithm}

\subsection{NewtonFool} \label{subsection:newtonfool}
\bbox{art.attacks.NewtonFool}{art/attacks/newtonfool.py}

NewtonFool~\citep{jang2017} is an untargeted attack that tries to decrease the probability $F_y(\boldsymbol{x})$ of the original class $y=C(\boldsymbol{x})$ by performing gradient descent.
The step size $\delta$ is determined adaptively in Equation~\eqref{alg:NewtonFool:delta}: when $F_y(\boldsymbol{x}_{\mbox{\tiny adv}})$ is larger than $1/K$ (recall $K$ is the number of classes in ${\cal Y}$), which is the case as long as $C(\boldsymbol{x}_{\mbox{\tiny adv}})=y$, then the second term will dominate; once $F_y(\boldsymbol{x}_{\mbox{\tiny adv}})$ approaches or falls below $1/K$, the first term will dominate. The tuning parameter $\eta$ determines how aggressively the gradient descent attempts to minimize the probability of class $y$.
The method is described in detail in Algorithm~\ref{alg:NewtonFool}.

\begin{algorithm}
  \caption{NewtonFool attack}
   \label{alg:NewtonFool}
\begin{algorithmic}[1]
\REQUIRE $\mbox{ }$ \\
$\boldsymbol{x}$: Input to be adversarially perturbed \\
$\eta$: Strength of adversarial perturbations \\
$i_{\mbox{\tiny max}}$: Maximum number of iterations
\STATE $y \leftarrow C(\boldsymbol{x})$, $\boldsymbol{x}_{\mbox{\tiny adv}} \leftarrow \boldsymbol{x}$, $i \leftarrow 0$
\WHILE{$i<i_{\mbox{\tiny max}}$}
\STATE Compute
\begin{eqnarray}
\delta &\leftarrow& \min \big\{ \eta \cdot \| \boldsymbol{x} \|_2 \cdot \|
\nabla F_y(\boldsymbol{x}_{\mbox{\tiny adv}}) \| , \, F_y(\boldsymbol{x}_{\mbox{\tiny adv}}) - 1/K \big\}, \label{alg:NewtonFool:delta} \\
\boldsymbol{d} &\leftarrow& - \frac{\delta \cdot \nabla F_y(\boldsymbol{x}_{\mbox{\tiny adv}})}{\| \nabla F_y(\boldsymbol{x}_{\mbox{\tiny adv}})\|_2^2} \nonumber
\end{eqnarray}
\STATE $\boldsymbol{x}_{\mbox{\tiny adv}} \leftarrow \mbox{clip}(\boldsymbol{x}_{\mbox{\tiny adv}} + \boldsymbol{d}, x_{\mbox{\tiny min}}, x_{\mbox{\tiny max}})$
\STATE $i \leftarrow i+1$
\ENDWHILE
\ENSURE  $\mbox{ }$ \\
Adversarial sample $\boldsymbol{x}_{\mbox{\tiny adv}}$.
\end{algorithmic}
\end{algorithm}

\subsection{Virtual Adversarial Method} \label{subsection:virtual_adversarial}
\bbox{art.attacks.VirtualAdversarialMethod}{art/attacks/virtual\_adversarial.py}

The Virtual Adversarial Method~\citep{miyato2015distributional} is not intended to create adversarial samples resulting in misclassification, but rather samples that, if included in the training set for adversarial training, result in local distributional smoothness of the trained model.
We use $N$ to denote the total number of components of classifier inputs and assume, for sake of convenience, that ${\cal X}={\mathbb R}^N$. By $\mbox{N}(\boldsymbol{0}, \boldsymbol{I_N})$ we denote a random sample of the $N$-dimensional standard normal distribution, and by $\boldsymbol{e}_i$ the $i$th standard basis vector of dimension $N$.
The key idea behind the algorithm is to construct a perturbation $\boldsymbol{d}$ with $\ell_2$ unit norm maximizing the Kullback-Leibler (KL) divergence $\mbox{KL}[F(\boldsymbol{x}) \| F(\boldsymbol{x}+\boldsymbol{d}) ]$.
In Algorithm~\ref{alg:VirtualAdversarial} this is done iteratively via gradient ascent along finite differences (line~\ref{alg:VirtualAdversarial:finite_differences}).
The final adversarial example $\boldsymbol{x}_{\mbox{\tiny adv}}$ is constructed by adding $\epsilon \cdot \boldsymbol{d}$ to the original input $\boldsymbol{x}$ (line~\ref{alg:VirtualAdversarial:x_adv}), where $\epsilon$ is the perturbation strength parameter provided by the user.

\begin{algorithm}
  \caption{Virtual Adversarial Method with finite differences}
   \label{alg:VirtualAdversarial}
\begin{algorithmic}[1]
\REQUIRE $\mbox{ }$ \\
$\boldsymbol{x}$: Input to be adversarially perturbed \\
$\epsilon$: Perturbation strength \\
$\xi$: Finite differences width\\
$i_{\mbox{\tiny max}}$: Maximum number of iterations
\STATE $\boldsymbol{d}\leftarrow \mbox{N}(\boldsymbol{0}, \boldsymbol{I_N})$
\STATE $\boldsymbol{d}\leftarrow  \boldsymbol{d}/\|\boldsymbol{d}\|_2$
\STATE $i \leftarrow 0$
\WHILE{$i < i_{\mbox{\tiny max}}$}
\STATE $\kappa_1 \leftarrow \mbox{KL}[F(\boldsymbol{x}) \| F(\boldsymbol{x}+\boldsymbol{d}) ]$
\STATE $\boldsymbol{d}_{\mbox{\tiny new}} \leftarrow \boldsymbol{d}$
\FOR{$i=1,2,\ldots,N$}
\STATE $\kappa_2 \leftarrow \mbox{KL}[F(\boldsymbol{x}) \| F(\boldsymbol{x}+\boldsymbol{d}+\xi\cdot\boldsymbol{e}_i) ]$
\STATE $\boldsymbol{d}_{\mbox{\tiny new}} \leftarrow \boldsymbol{d}_{\mbox{\tiny new}} + (\kappa_2-\kappa_1)/\xi\cdot\boldsymbol{e}_i$ \label{alg:VirtualAdversarial:finite_differences}
\ENDFOR
\STATE $\boldsymbol{d} \leftarrow \boldsymbol{d}_{\mbox{\tiny new}}$
\STATE $\boldsymbol{d}\leftarrow  \boldsymbol{d}/\|\boldsymbol{d}\|_2$
\ENDWHILE
\STATE  $\boldsymbol{x}_{\mbox{\tiny adv}} \leftarrow \mbox{clip}(\boldsymbol{x} + \epsilon \cdot \boldsymbol{d}, x_{\mbox{\tiny min}}, x_{\mbox{\tiny max}})$ \label{alg:VirtualAdversarial:x_adv}
\ENSURE  $\mbox{ }$ \\
Adversarial sample $\boldsymbol{x}_{\mbox{\tiny adv}}$.
\end{algorithmic}
\end{algorithm}

\subsection{Spatial Transformation Attack} \label{subsection:spatial_transformation}
\bbox{art.attacks.SpatialTransformation}{art/attacks/spatial\_transformation.py}

The Spatial Transformation Attack proposed by \citep{engstrom_spatial2017} performs a combination of exactly one translation and one rotation of the input image in order to generate an adversarial sample. The same translation and rotation parameters are used for the entire batch of inputs (i.e.~in some sense the adversarial perturbation is ``universal''), and the optimal combination of translation and rotation parameters for the entire batch is found via grid search.

\subsection{Elastic Net (EAD) Attack} \label{subsection:ead}
\bbox{art.attacks.ElasticNet}{art/attacks/elastic\_net.py}

The Elastic Net (EAD) Attack proposed by \citep{chen2017ead} is a modification of Carlini and Wagner's attack (cf.~Section \ref{subsection:CW}) which aims at controlling the $\ell_1$ norm of adversarial perturbations.

\subsection{ZOO Attack} \label{subsection:zoo}
\bbox{art.attacks.ZooAttack}{art/attacks/zoo.py}

The Zeroth-Order-Optimization (ZOO) Attack proposed by \citep{chen2017zoo} is a black-box version of Carlini and Wagner's attack (cf.~Section \ref{subsection:CW}) which relies on queries of the classifier's output probabilities.

\subsection{Boundary Attack} \label{subsection:boundary_attack}
\bbox{art.attacks.BoundaryAttack}{art/attacks/boundary.py}

The Boundary Attack~\citep{brendel2017} is a black-box attack which only requires queries of the output class, not of the logit or of output probabilities.

\subsection{Adversarial Patch} \label{subsection:adversarial_patch}
\bbox{art.attacks.AdversarialPatch}{art/attacks/adversarial\_patch.py}

An Adversarial Patch~\citep{brown2017patch} is an adversarial sample designed in such a way that, when printed out and inserted into a natural scene, pictures taken of that scene will be misclassified as a given target class.

\subsection{Decision Tree Attack} \label{subsection:decision_tree_attack}
\bbox{art.attacks.DecisionTreeAttack}{art/attacks/decision\_tree\_attack.py}

This implementation of Papernot's attack on decision trees \citep{papernot2016transferability} follows Algorithm 2 and communication with the authors.

\subsection{High Confidence Low Uncertainty (HCLU) Attack} \label{subsection:hclu_attack}
\bbox{art.attacks.HighConfidenceLowUncertainty}{art/attacks/hclu.py}

Implementation of the High-Confidence-Low-Uncertainty (HCLU) adversarial example formulation \cite{Grosse2018}.

\subsection{HopSkipJump Attack} \label{subsection:hop_skip_jump_attack}
\bbox{art.attacks.HopSkipJump}{art/attacks/hop\_skip\_jump.py}

Implementation the HopSkipJump attack \citep{Chen2019HopSkipJump}. This is a black-box attack that only requires class predictions. It is an advanced version of the Boundary attack.

\section{Defences} \label{sec:defences}
\bbox{art.defences}{}

There is an increasing number of methods for defending against evasion attacks which can roughly be categorized into:
\begin{description}
  \item[Model hardening] refers to techniques resulting in a new classifier with better robustness properties than the original one with respect to some given metrics.
  \item[Data preprocessing] techniques achieve higher robustness by using transformations of the classifier inputs and labels, respectively, at test and/or training time.
  \item[Runtime detection] of adversarial samples by extending the original classifier with a detector in order to check whether a given input is adversarial or not.
\end{description}

The following defences are currently implemented in ART:
\begin{itemize}

\item Adversarial Training, Section~\ref{subsec:Adversarial_training}
\item Feature Squeezing, Section~\ref{subsec:Feature_squeezing}
\item Label Smoothing, Section~\ref{subsec:Label_smoothing}
\item Spatial Smoothing, Section~\ref{subsec:Spatial_smoothing}
\item JPEG Compression, Section~\ref{sec:jpeg}
\item Thermometer Encoding, Section~\ref{sec:thermometer}
\item Total Variance Minimization, Section~\ref{sec:tvm}
\item Gaussian Augmentation, Section~\ref{sec:gaussian}
\item Pixel Defend, Section~\ref{sec:pixel_defend}

\end{itemize}

\subsection{Adversarial Training} \label{subsec:Adversarial_training}
\bbox{art.defences.AdversarialTrainer}{art/defences/adversarial\_trainer.py}

The idea of adversarial training~\citep{goodfellow2014fgsm} is to improve the robustness of the classifier $C(\boldsymbol{x})$ by including adversarial samples in the training set.
A special case of adversarial training is virtual adversarial training~\citep{miyato2017virtual} where the adversarial samples are generated by the Virtual Adversarial Method (see Section~\ref{subsection:virtual_adversarial}).
 Given a set of attacks and classifiers $\{(\rho_1, C_1),\ldots,(\rho_m, C_m)\}$ and the original training set $\{(\boldsymbol{x}_1,y_1),\ldots, (\boldsymbol{x}_n, y_n) \}$, the ART implementation statically enhances the training set with the adversarial samples $(\rho_i(\boldsymbol{x}_j), C_i(\rho_i(\boldsymbol{x}_j)))$ for $i=1,\ldots,m$ and $j=1,\ldots,n$.
 As a consequence, if $n$ attacks are applied to all original training samples, the resulting augmented training set will have a size of $m\cdot n$.
 The augmented data is then used to train a hardened classifier $C(\boldsymbol{x})$.
 This idea of combining adversarial samples from multiple attacks for adversarial training has also been explored in~\citep{tramer2017ensemble}.

If multiple attacks are specified, they are rotated for each batch. If the specified attacks have as target a different model that the one being trained, then the attack is transferred.
A ratio parameter determines how many of the clean samples in each batch are replaced with their adversarial counterpart.

\paragraph{Implementation details}
The {\tt AdversarialTrainer} class has the following public functions:
\begin{itemize}
\item {\tt \_\_init\_\_(classifier, attacks, ratio=.5),}

The parameters of the constructor are the following:
  \begin{itemize}
    \item {\tt classifier}: The classifier $C(\boldsymbol{x})$ to be hardened by adversarial training.
    \item {\tt attacks}: Attack or list of attacks to be used for adversarial training. These are instances of the \texttt{Attack} class.
    \item {\tt ratio}: The proportion of samples in each batch to be replaced with their adversarial counterparts.
                      Setting this value to 1 allows to train only on adversarial samples.
  \end{itemize}
  Each instance of the class {\tt Attack} (corresponding to $\rho_i(\boldsymbol{x})$ in the notation above), has an attribute {\tt classifier} with the classifier the attack aims at, corresponding to $C_i(\boldsymbol{x})$ in the notation above.
\item {\tt fit(x, y, batch\_size=128, nb\_epochs=20) -> None}

Method for training the hardened classifier. {\tt x} and {\tt y} are the original training inputs and labels, respectively.
The method applies the adversarial attacks specified in the {\tt attacks} parameter of \texttt{\_\_init\_\_} to obtain the enhanced training set as explained above.
Then the {\tt fit} function of the classifier specified in the class attribute {\tt classifier} is called with the enhanced training set as input.
The hardened classifier is trained with the given batch size for as many epochs as specified.
After calling this function, the {\tt classifier} attribute of the class contains the hardened classifier $C(\boldsymbol{x})$.

\item {\tt fit\_generator(generator, nb\_epochs=20)}

Train a model adversarially using a data generator for the number of epochs that was specified.
Adversarial training is performed using the configuration specified in the \texttt{\_\_init\_\_} function.
Here, \texttt{generator} is an instance following the \texttt{DataGenerator} interface (see Section~\ref{sec:data_gen} for details), which creates a wrapper on top of custom or framework-specific data feeders.
Like in the case of \texttt{fit}, after calling this function, the {\tt classifier} attribute of the class contains the hardened classifier $C(\boldsymbol{x})$.

\item {\tt predict(x, $\ast\ast$kwargs) -> np.ndarray}

Calls the {\tt predict} function of the hardened classifier $C(\boldsymbol{x})$, passing on the dictionary of arguments \texttt{$\ast\ast$kwargs)}.
\end{itemize}




\subsection{The Data \texttt{Preprocessor} Base class} \label{subsec:Preprocessing}
\bbox{art.defences.Preprocessor}{art/defences/preprocessor.py}

The abstract class {\tt Preprocessor} provides a unified interface for all data preprocessing transformations to be used as defences.

The public interface of the {\tt Preprocessor} grants access to the following properties:
\begin{itemize}
  \item {\tt is\_fitted}: inidcating whether the transformations have been fitted (where applicable).
  \item \texttt{apply\_fit}: indicating whether or not the defence should be applied at training time.
  \item \texttt{apply\_predict}: indicating whether or not the defence should be applied at test time.

\end{itemize}

The class has the following public methods:
\begin{itemize}

\item {\tt fit(x, y=None, $\ast\ast$kwargs)}: Fit the transformations with the given data and parameters (where applicable).
\item {\tt \_\_call\_\_(x, y=None)}: Applies the transformations to the provided labels and/or inputs and returns the transformed data.

\item {\tt estimate\_gradient(x, grad)}: Provide the gradient of the defence (or an approximation thereof in case the defence is not differentiable) for a backward pass. Here \texttt{x} is the input data for which the gradient is computed, and \texttt{grad} is the gradient value through the backward pass so far.           

\end{itemize}

\subsection{Feature Squeezing} \label{subsec:Feature_squeezing}
\bbox{art.defences.FeatureSqueezing}{art/defences/feature\_squeezing.py}

Feature squeezing~\citep{xu2017feature_squeeze} reduces the precision of the components of $\boldsymbol{x}$ by encoding them on a smaller number of bits.
In the case of images, one can think of feature squeezing as reducing the common 8-bit pixel values to $b$ bits where $b<8$.
Formally, assuming that the components of $\boldsymbol{x}$ all lie within $[0,1]$ (which would have been obtained, e.g., by dividing 8-bit pixel values by 255) and given the desired {\bf bit depth} $b$, feature squeezing applies the following transformation component-wise:
\begin{eqnarray}
\boldsymbol{x} &\leftarrow& \lfloor \boldsymbol{x} \cdot (2^b -1) \rfloor / (2^b -1) \label{eq:Feature_squeezing}.
\end{eqnarray}

Since feature squeezing does not require any model fitting for the data transformations, {\tt is\_fitted} always returns {\tt True}, and the {\tt fit} function does not have any effect.

Note that any instance of the {\tt Classifier} class which has feature squeezing as one of their defences will automatically apply this operation when the {\tt fit} or {\tt predict} functions are called.

%
%
%
%
%

\subsection{Label Smoothing} \label{subsec:Label_smoothing}
\bbox{art.defences.LabelSmoothing}{art/defences/label\_smoothing.py}

Label smoothing~\citep{hazan2016perturbation} modifies the labels $y$ during model training: instead of using one-hot encoding, where $y$ is represented by the standard basis vector $\boldsymbol{y}=\boldsymbol{e}_y$, the following representation is used:
\begin{eqnarray*}
\boldsymbol{y}_i &\leftarrow& \left\{
\begin{array}{ll}
y_{\mbox{\tiny max}} & \mbox{if $i=y$} \\
 (1-y_{\mbox{\tiny max}})/(K-1) & \mbox{otherwise,}
\end{array}
\right.
\end{eqnarray*}
where $y_{\mbox{\tiny max}}\in[0,1]$ is a parameter specified by the user.
The label representation $\boldsymbol{y}$ is thus ``smoothed'' in the sense that the difference between its maximum and minimum components is reduced and its entropy increased. The motivation behind this approach is that it might help reducing gradients that an adversary could exploit in the construction of adversarial samples.

Note that any instance of the {\tt Classifier} class which has label smoothing activated in the defences will automatically apply it when the {\tt fit} function is called.

%
%
%
%
%

\subsection{Spatial Smoothing} \label{subsec:Spatial_smoothing}
\bbox{art.defences.SpatialSmoothing}{art/defences/spatial\_smoothing.py}

Spatial smoothing~\citep{xu2017feature_squeeze} is a defence specifically designed for images.
It attempts to filter out adversarial signals using local spatial smoothing.
Let $x_{ijk}$ denote the components of $\boldsymbol{x}$.
Recall that $i$ indexes width, $j$ height and $k$ the color channel. Given a {\bf window size} $w$, the component $x_{ijk}$ is replaced by a median-filtered version:
\begin{eqnarray*}
x_{ijk} \leftarrow \mbox{median}\big\{x_{i'j'k}: \  i-\lfloor w/2 \rfloor \leq i' \leq  i+\lceil w/2 \rceil -1,\
j-\lfloor w/2 \rfloor \leq j' \leq  j+\lceil w/2 \rceil -1 \big\},
\end{eqnarray*}
where features at the borders are reflected where needed:
\begin{itemize}
\item $x_{ijk}=x_{(1-i),j,k}$ for $i=0,-1,\ldots$
\item $x_{(k_1+i),j,k}=x_{(k_1+1-i),j,k}$ for $i=1,2,\ldots$
\end{itemize}
and analogously for $j$. Note that the local spatial smoothing is applied separately in each color channel $k$.

Same as for the previous two defences, {\tt is\_fitted} is always {\tt True}, and the {\tt fit} function does not have any effect.

Note that any instance of the {\tt Classifier} class which has spatial smoothing as one of their defences will automatically apply it when the {\tt predict} function is called.

%
%
%
%
%

\subsection{JPEG Compression} \label{sec:jpeg}
\bbox{art.defences.JpegCompression}{art/defences/jpeg\_compression.py}

JPEG compression~\citep{dziugaite2016,das2017} can work as an effective preprocessing step in the classification pipeline to counter adversarial attacks and dramatically reduce their effect.
An important component of JPEG compression is its ability to remove high frequency signal components, inside square blocks of an image.
Such an operation is equivalent to selective blurring of the image, helping remove additive perturbations.

\subsection{Thermometer Encoding} \label{sec:thermometer}
\bbox{art.defences.ThermometerEncoding}{art/defences/thermometer\_encoding.py}

Thermometer encoding~\citep{buckman2018} is a preprocessing method for input discretization that encodes each feature as a fixed-size binary vector.
To this end, the input domain is first evenly divided into $b$ distinct buckets.
$b$ represents the number of bits that will be used for encoding each feature.
The encoded value per feature corresponds to a number of ones equal to the index of the bucket containing the original value.
The ones are filled from the back of the vector.
As the representation is fixed-size, the rest of the vector is filled with zeros.
Note that the thermometer encoding preserves pairwise ordering information.

Applying this method does not require training, thus the \texttt{fit} function is a dummy.

\subsection{Total Variance Minimization} \label{sec:tvm}
\bbox{art.defences.TotalVarMin}{art/defences/variance\_minimization.py}

Total variance minimization~\citep{guo2018} randomly selects a small set of pixels, and reconstructs the simplest image that is consistent with the selected pixels.
The reconstructed image does not contain the adversarial perturbations because these perturbations tend to be small and localized.

Specifically, a small number of pixels is sampled using a Bernoulli random variable $X(i, j, k)$ for each pixel location $(i, j, k)$.
Next, total variation minimization is used to construct an image $\boldsymbol{z}$ that is similar to the input image $\boldsymbol{x}$ for the selected set of pixels, whilst also being “simple” in terms of total variation by solving:
\begin{equation}
  \min_{\boldsymbol{z}} || (1-X) \odot (\boldsymbol{z} - \boldsymbol{x})||_2 + \lambda \cdot \mbox{TV}_p (\boldsymbol{z}), \label{eq:tvm}
\end{equation}
where $\mbox{TV}_p(\boldsymbol{z})$ is the $L_p$ total variation of $\boldsymbol{z}$:
\[
  \mbox{TV}_p(\boldsymbol{z}) = \sum_{k=1}^K \left[ \sum_{i=2}^N ||\boldsymbol{z}(i,:,k) - \boldsymbol{z}(i-1,:,k)||_p + \sum_{j=2}^N ||\boldsymbol{z}(:,j,k) - \boldsymbol{z}(:,j-1,k)||_p \right].
\]

The objective function in Equation~\eqref{eq:tvm} is convex in $\boldsymbol{z}$.
In the original method, the norm $p$ is set to $L_2$, and the authors employ a special-purpose solver based on the split Bregman method to perform total variance minimization efficiently.

\subsection{Gaussian Data Augmentation} \label{sec:gaussian}
\bbox{art.defences.GaussianAugmentation}{art/defences/gaussian\_augmentation.py}

Gaussian data augmentation~\citep{zantedeschi2017} is a standard data augmentation technique in computer vision that has also been used to improve the robustness of a model to adversarial attacks.
This method augments a dataset with copies of the original samples to which Gaussian noise has been added.
It can also be used to add Gaussian noise to an existing sample without augmentation.
An important advantage of this defence is its independence from the attack strategy.
Its usage is mostly intended for the augmentation of the training set.
Applying this method does not require training, thus the \texttt{fit} function is a dummy.
Notice that the labels \texttt{y} are not required, as the method does not apply any preprocessing to them.


%
%

\subsection{Pixel Defence} \label{sec:pixel_defend}
\bbox{art.defences.PixelDefend}{art/defences/pixel\_defend.py}

PixelDefend~\citep{song2018} uses a CNN to move adversarially perturbed inputs back to the data manifold before feeding them to the original classifier.
Applying this method does not require training, thus the \texttt{fit} function is a dummy.
Notice that the labels \texttt{y} are not required, as the method does not apply any preprocessing to them.

\section{Evasion Detection}
\bbox{art.detection}{}

This module fills the purpose of providing runtime detection methods for adversarial samples.
Currently, the library implements two types of detectors:

\begin{itemize}

\item BinaryInputDetector, Section~\ref{sec:binary_detector}

\item BinaryActivationDetector, Section~\ref{sec:binary_activation_detector}
  
\item SubsetScanningDetector, Section~\ref{sec:subsetscanning_detector}

\end{itemize}

The previous methods implement the API provided in the \texttt{Detector} base class, that we describe in the following.

\subsection{The \texttt{Detector} Base Class} \label{sec:detection}
\bbox{art.detection.Detector}{art/detection/detector.py}

The \texttt{Detector} abstract class provides a unified interface for all runtime adversarial detection methods.
The class has the following public methods:
\begin{itemize}
\item {\tt fit(x, y=None, $\ast\ast$kwargs) -> None}

Fit the detector with the given data and parameters (where applicable).

\item {\tt \_\_call\_\_(x) -> np.ndarray}

Applies detection to the provided inputs and returns binary decisions for each input sample.

\item {\tt is\_fitted}

Check whether the detector has been fitted (where applicable).
\end{itemize}

\subsection{Binary Detector Based on Inputs} \label{sec:binary_detector}
\bbox{art.detection.BinaryInputDetector}{art/detection/detector.py}

This method builds a binary classifier, where the labels represent the fact the a given input is adversarial (label 1) or not (label 0).
The detector is fit with a mix of clean and adversarial data; following this step, it is ready to detect adversarial inputs.

\subsection{Binary Detector Based on Activations} \label{sec:binary_activation_detector}
\bbox{art.detection.BinaryActivationDetector}{art/detection/detector.py}

This method builds a binary classifier, where the labels represent the fact the a given input is adversarial (label 1) or not (label 0).
The activations detector is different from the previous one in that it uses as inputs for training the values of the activations of a different classifier.
Only activations from one layer that is specified are used.

\subsection{Fast Generalized Subset Scan Based detector} \label{sec:subsetscanning_detector}
\bbox{art.detection.subsetscanning.SubsetScanningDetector \\
.}{art/detection/subsetscanning/detector.py}

Fast generalized subset scan based detector \citep{McFowland2013SubsetScan} to detect anomalous patterns
in general categorical data sets.
        
\section{Poisoning Detection} \label{sec:poison}
\bbox{art.poison\_detection}{}

Data used to train machine learning models are often collected from potentially untrustworthy sources.
This is particularly true for crowdsourced data (e.g. Amazon Mechanical Turk), social media data, and data collected from user behavior (e.g. customer satisfaction ratings, purchasing history, user traffic).
Adversaries can craft inputs to modify the decision boundaries of machine learning models to misclassify inputs or to reduce model performance.

As part of targeted misclassification attacks, recent work has shown that adversaries can generate “backdoors" or “trojans” into machine learning models by inserting malicious data into the training set \cite{badnets}. The resulting model performs very well on the intended training and testing data, but behaves badly on specific attacker-chosen inputs. As an example, it has been demonstrated that a network trained to identify street signs has strong performance on standard inputs, but identifies stop signs as speed limit signs when a special sticker is added to the stop sign. This backdoor provides adversaries a method for ensuring that any stop sign is misclassified simply by placing a sticker on it. Unlike adversarial samples that require specific, complex noise to be added to an image, backdoor triggers can be quite simple and can be easily applied to an image - as easily as adding a sticker to a sign or modifying a pixel.

ART provides filtering defences to detect poison this type of attacks.

\subsection{The \texttt{PoisonFilteringDefence} Base Class}
\bbox{\small{art.poison\_detection.PoisonFilteringDefence}}{\small{art/poison\_detection/poison\_filtering\_defence.py}}

The abstract class \texttt{PoisonFilteringDefence} defines an interface for the detection of poison when the training data is available.
This class takes a model and its corresponding training data and identifies the set of data points that are suspected of being poisonous.

\subsection{Poisoning Filter Using Activation Clustering Against Backdoor Attacks}
\bbox{\small{art.poison\_detection.ActivationDefence}}{\small{art/poison\_detection/activation\_defence.py}}

\begin{figure*}
\centering
\begin{subfigure}[m]{.4\textwidth}
\caption{}
\label{fig:poisonous_activations_RT}
\includegraphics[width=\textwidth]{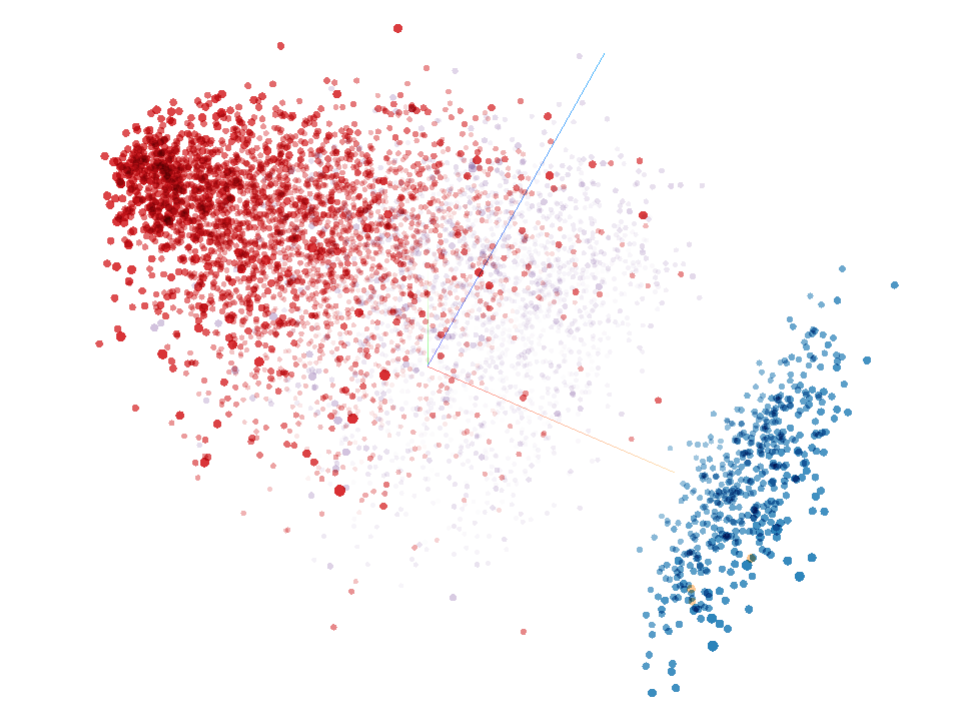}
\end{subfigure}~~~~~~~~~~~~~~~~~~~~~
\begin{subfigure}[m]{.4\textwidth}
\caption{}
\label{fig:clean_activations_RT}
\includegraphics[width=\textwidth]{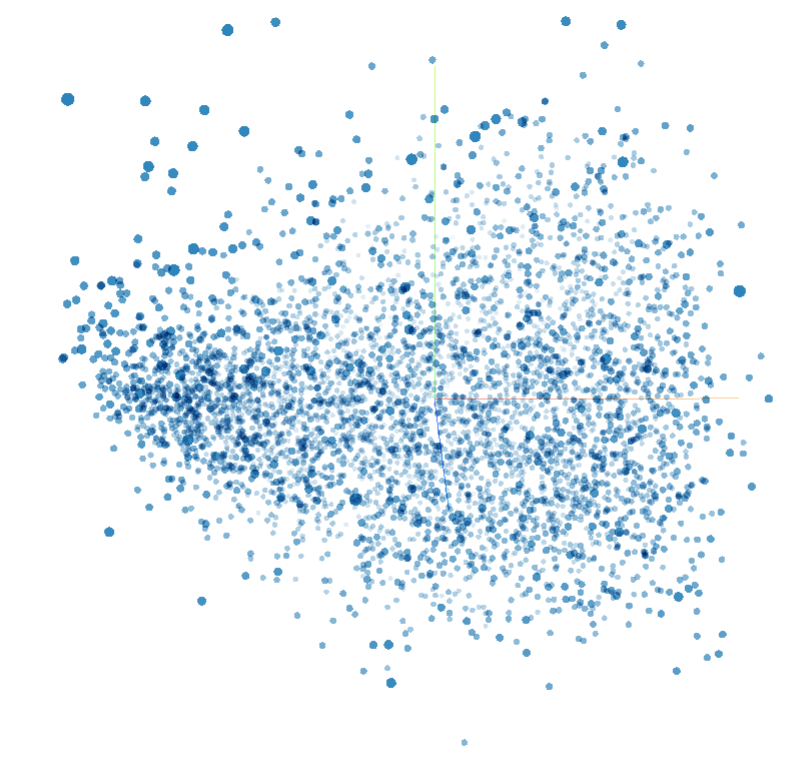}
\end{subfigure}
\caption{Activations of the last hidden layer projected onto the first 3 principle components.
(a) Activations of the \emph{poisoned} class (d) Activations of the \emph{unpoisoned} class.}
\label{fig:ac-clustering}
\end{figure*}

The Activation Clustering \cite{activation-defence} defence detects poisonous data crafted to insert backdoors into neural networks.
The intuition behind this method is that while backdoor and target samples receive the same classification by the poisoned network, the reason why they receive this classification is different. In the case of standard samples from the target class, the network identifies features in the input that it has learned correspond to the target class. In the case of backdoor samples, it identifies features associated with the source class \emph{and} the backdoor trigger, which causes it to classify the input as the target class. This difference in mechanism should be evident in the network activations, which represent how the network made its ``decisions''.
This intuition is verified in Figure \ref{fig:ac-clustering}, which shows activations of the last hidden neural network layer for different scenarios.
Figure \ref{fig:poisonous_activations_RT} shows the activations of the poisoned negative class for a neural network trained on the Rotten Tomatoes movie reviews dataset. It is easy to see that the activations of the poisonous and legitimate data break out into two distinct clusters. In contrast, Figure \ref{fig:clean_activations_RT} displays the activations of the positive class, which was not targeted with poison. Here we see that the activations do not break out into two distinguishable clusters.

The Activation Clustering defence is formalized in Algorithm \ref{alg:AC}.
As input, the algorithm takes the data used to train the provided classifier.
Each sample in the untrusted training set is classified and the activations of the last hidden layer are retained.
These activations are segmented according to their labels. For each activation segment,
the dimensionality is reduced, and then a clustering algorithm is applied.
To cluster the activations, activations are reshaped into a 1D vector and a dimensionality reduction is performed using Independent Component Analysis (ICA), which was found to be more effective than Principle Component Analysis (PCA). Dimensionality reduction before clustering is necessary to avoid known issues with clustering on very high dimensional data \citep{aggarwal2001surprising, domingos2012few}.

$k$-means with $k=2$ was found to be highly effective at separating the poisonous from legitimate activations. Other clustering methods were tested including DBSCAN, Gaussian Mixture Models, and Affinity Propagation, but $k$-means was superior in terms of speed and accuracy. However, $k$-means will separate the activations into two clusters, regardless of whether poison is present or not. Thus, it is still necessary to determine which, if any, of the clusters corresponds to poisonous data.
Multiple clustering analysis techniques are implemented in this release including relative cluster size, inter-cluster distance and cluster cohesiveness.

\begin{figure}
\centering
\includegraphics[width=0.9\textwidth]{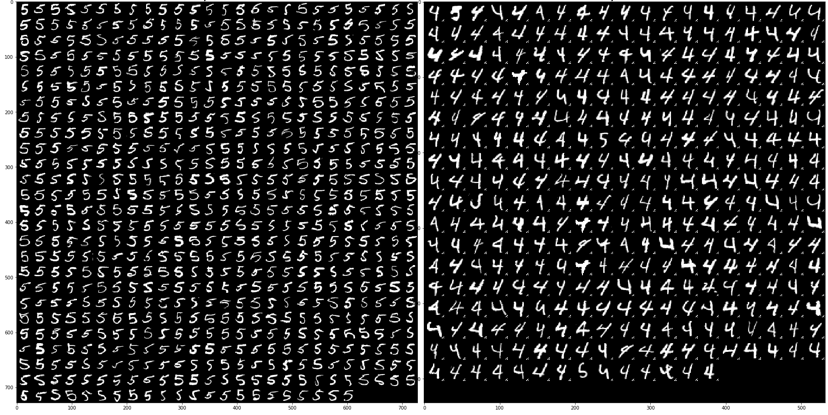}
\caption{Library visualization of clusters for class five produce by the Activation Clustering defence (see code example below).
Sprite in the left contains samples in the first cluster, while sprite in the right contains samples in the second cluster.
It is easy to see the second cluster contains poison data.}
\label{fig:sample-output}
\end{figure}

Analysts can also manually review the results using the visualization method included in this release.
An example visualization for MNIST dataset is presented in Figure \ref{fig:sample-output}.
Here, the library generated a sprite containing the images in each of the clusters generated
by the Activation Clustering defence for digit five.
It is easy to see that one of the clusters contains a backdoor (fours with a small dot in the bottom right).
After poisonous data is identified, the model needs to be repaired accordingly.

\begin{algorithm} [htb]
	\caption{Backdoor Detection Activation Clustering Algorithm}
	{\bfseries Inputs:} $D_p$ := Untrusted training dataset with class labels $\{1, ..., n\}$ 	 and
	$F_{\Theta_P}$:= Neural network trained with $D_p$
	\label{alg:AC}
\begin{algorithmic} [1]
    	\STATE Initialize $A$; $A[i]$ holds activations for all $s_i \in D_p$ such that $F_{\Theta_P}(s_i) = i$
	\FORALL{$s\in D_p$}
		\STATE $A_s \leftarrow$ activations of last hidden layer of $F_{\Theta_P}$ flattened into a single 1D vector
		\STATE Append $A_s$ to $A[F_{\Theta_P}(s)]$
	\ENDFOR
	\FORALL{$i=0$ {\bfseries to} $n$}
		\STATE red = reduceDimensions$(A[i])$
		\STATE clusters = clusteringMethod$(red)$ \label{ln:clustering}
		\STATE analyzeForPoison(clusters) \label{ln:cluster-analysis}
	\ENDFOR
\end{algorithmic}
\end{algorithm}

\section{Metrics} \label{sec:metrics}
\bbox{art.metrics}{}

For assessing the {\bf robustness} of a classifier against adversarial attacks, possible {\bf metrics} are: the average minimal perturbation that is required to get an input misclassified; the average sensitivity of the model's loss function with respect to changes in the inputs; the average sensitivity of the model's logits with respect to changes in the inputs, e.g.~based on Lipschitz constants in the neighborhood of sample inputs.
The {\tt art.metrics} module implements several metrics to assess the robustness respectively vulnerability of a given classifier, either generally or with respect to specific attacks:
\begin{itemize}
\item Empirical robustness, Section~\ref{subsec:Empirical_robustness}
\item Loss sensitivity, Section~\ref{subsec:Loss_sensitivity}
\item CLEVER score, Section~\ref{subsec:CLEVER}
\item Clique method robustness verification for decision tree ensembles, Section~\ref{subsec:clique_method}
\end{itemize}

\subsection{Empirical Robustness} \label{subsec:Empirical_robustness}
\bbox{art.metrics.metrics.empirical\_robustness}{art/metrics/metrics.py}

Empirical robustness assesses the robustness of a given classifier with respect to a specific attack and test data set. It is equivalent to the average minimal perturbation that the attacker needs to introduce for a successful attack, as introduced in~\citep{DBLP:journals/corr/Moosavi-Dezfooli15}.
Given a trained classifier $C(\boldsymbol{x})$, an untargeted attack $\rho(\boldsymbol{x})$ and test data samples
$\boldsymbol{X}=(\boldsymbol{x}_1, \ldots, \boldsymbol{x}_n)$, let $I$ be the subset of indices $i\in \{1,\ldots,n\}$ for which
$C(\rho(\boldsymbol{x}_i))\neq C(\boldsymbol{x}_i)$, i.e.~for which the attack was successful.
Then the empirical robustness (ER) is defined as:
\begin{eqnarray*}
\mbox{ER}(C, \rho, \boldsymbol{X}) &=& \frac{1}{|I|} \sum_{i\in I}
\frac{\| \rho(\boldsymbol{x}_i)-\boldsymbol{x}_i \|_p}{ \|\boldsymbol{x}_i\|_p}.
\end{eqnarray*}
Here $p$ is the norm used in the creation of the adversarial samples (if applicable); the default value is $p=2$.

\subsection{Loss Sensitivity} \label{subsec:Loss_sensitivity}
\bbox{art.metrics.loss\_sensitivity}{art/metrics/metrics.py}

Local loss sensitivity aims to quantify the smoothness of a model by estimating its Lipschitz continuity constant, which measures the largest variation of a function under a small change in its input: the smaller the value, the smoother the function.
This measure is estimated based on the gradients of the classifier logits, as considered e.g.\ in~\citep{krueger2017memory}.
It is thus an attack-independent measure offering insight on the properties of the model.

Given a classifier $C(\boldsymbol{x})$ and a test set $\boldsymbol{X}=(\boldsymbol{x}_1,\ldots,\boldsymbol{x}_n)$, the loss sensitivity (LS) is defined as follows:
\begin{eqnarray*}
\mbox{LS}(C, \boldsymbol{X}, \mathbf{y}) &=& \frac{1}{n} \sum_{i=1}^n \big\| \nabla \mathcal{L}(\boldsymbol{x}_i, y_i) \big\|_2.
\end{eqnarray*}

\subsection{CLEVER} \label{subsec:CLEVER}
\bbox{art.metrics.clever\_u \\
art.metrics.clever\_t}{art/metrics/metrics.py}

The Cross Lipschitz Extreme Value for nEtwork Robustness metric (CLEVER)~\citep{weng2018} estimates, for a given input $\boldsymbol{x}$ and $\ell_p$ norm, a lower bound $\gamma$ for the minimal perturbation that is required to change the classification of $\boldsymbol{x}$, i.e.~$\| \boldsymbol{x}-\boldsymbol{x}' \|_p < \gamma$ implies $C(\boldsymbol{x})=C(\boldsymbol{x}')$ (see Corollary 3.2.2 in~\citep{weng2018}).
The derivation of $\gamma$ is based on a Lipschitz constant for the gradients of the classifier's logits in an $\ell_p$-ball with radius $R$ around $\boldsymbol{x}$.
Since in general there is no closed-form expression or upper bound for this constant\footnote[3]{Closed-form expressions for special types of classifier are derived in \citep{hein2017}, which is the first work considering Lipschitz bounds for deriving formal guarantees of classifiers robustness against adversarial inputs.}, the CLEVER algorithm uses an estimate based on extreme value theory.

Algorithm~\ref{alg:CLEVER} outlines the calculation of the CLEVER metric for targeted attacks: given a target class $y$, the score $\gamma$ is constructed such that $C(\boldsymbol{x}') \neq y$ as long as $\| \boldsymbol{x}-\boldsymbol{x}' \|_p < \gamma$.
Below we discuss how to adapt the algorithm to the untargeted case.
A key step in the algorithm is outlined in lines~\ref{alg:CLEVER:sample1}-\ref{alg:CLEVER:sample2} where the norm of gradient differences is evaluated at points randomly sampled from the uniform distribution on the $\ell_p$-ball with radius $R$ around $\boldsymbol{x}$.
The set $S_{\mbox{\tiny max}}$ collects the maximum norms from $n_{\mbox{\tiny batch}}$ batches of size $n_{\mbox{\tiny sample}}$ each (line~\ref{alg:CLEVER:max_values}), and $\hat{\mu}$ is the MLE of the location parameter $\mu$ of a reverse Weibull distribution given the realizations in $S$.
The final CLEVER score is calculated in line~\ref{alg:CLEVER:score_calc}.
Note that it is bounded by the radius $R$ of the $\ell_p$-ball, as for any larger perturbations the estimated Lipschitz constant might not apply.

The CLEVER score for {\bf untargeted attacks} is simply obtained by taking the minimum CLEVER score for targeted attacks over all classes $y\in{\cal Y}$ with $y\neq C(\boldsymbol{x})$. The method {\tt clever\_u} for calculating the CLEVER score of untargeted attacks has the same signature as {\tt clever\_t} except that it does not take the {\tt target\_class} parameter.

\begin{algorithm}
  \caption{CLEVER score for targeted attack}
   \label{alg:CLEVER}
\begin{algorithmic}[1]
\REQUIRE $\mbox{ }$ \\
$\boldsymbol{x}$: Input for which to calculate the CLEVER score \\
$y$: Target class \\
$n_{\mbox{\tiny batch}}$: Number of batches over each of which the maximum gradient is computed \\
$n_{\mbox{\tiny sample}}$: Number of samples per batch \\
$p$: Perturbation norm \\
$R$: Maximum $\ell_p$ norm of perturbations considered in the CLEVER score
\STATE $S_{\mbox{\tiny max}} \leftarrow \{\}$
\STATE $q \leftarrow p/(p-1)$
\FOR{$i=1,\ldots,n_{\mbox{\tiny batch}}$}
\FOR{$j=1,\ldots,n_{\mbox{\tiny sample}}$}
\STATE $\boldsymbol{x}_{j} \leftarrow B_p(\boldsymbol{x}, R)$ \label{alg:CLEVER:sample1}
\STATE $b_{j} \leftarrow \| \nabla Z_{C(\boldsymbol{x})}(\boldsymbol{x}_{i}) - \nabla Z_{y}(\boldsymbol{x}_{j}) \|_q$  \label{alg:CLEVER:sample2}
\ENDFOR
\STATE $S_{\mbox{\tiny max}} \leftarrow S_{\mbox{\tiny max}} \cup \{ \max_j b_{j}\}$ \label{alg:CLEVER:max_values}
\ENDFOR
\STATE $\hat{\mu} \leftarrow \mbox{MLE}_{\mbox{\tiny Weibull}}(S_{\mbox{\tiny max}})$ \label{alg:CLEVER:MLE}
\STATE $\gamma \leftarrow \min\{  (Z_{C(\boldsymbol{x})}(\boldsymbol{x}) -  Z_{y}(\boldsymbol{x}))/\hat{\mu},\,R \}$ \label{alg:CLEVER:score_calc}
\ENSURE  $\mbox{ }$ \\
CLEVER score $\gamma$.
\end{algorithmic}
\end{algorithm}

\subsection{Clique Method Robustness Verification for Decision Tree Ensembles} \label{subsec:clique_method}
\bbox{art.metrics.RobustnessVerificationTreeModelsCliqueMethod \\
.}{art/metrics/verification\_decisions\_trees.py}

Robustness verification for decision-tree ensemble models like Gradinet Boosted Decision Trees (XGBoost, LightGBM, Scikit-learn), Random Forest (Scikit-learn), or Extra Trees (Scikit-learn) based on clique method \citep{Chen2019Clique}.

\section{Data Generators} \label{sec:data_gen}
\bbox{art.data\_generators.DataGenerator\\
art.data\_generators.KerasDataGenerator\\
art.data\_generators.MXDataGenerator\\
art.data\_generators.PyTorchDataGenerator}{art/data\_generators.py}

The \texttt{DataGenerator} interface provides a standardized way for incorporating data loader and generators.
These are most useful when working with large datasets that do not fit into memory and that are loaded one batch at a time.
Another use case for this interface is when performing data augmentation on the fly for each batch.
Users can define their own generators performing loading or data augmentation, that can be subsequently used with the \texttt{fit\_generator} functions available in the \texttt{Classifier} and in the \texttt{AdversarialTrainer}.
The \texttt{DataGenerator} interface consists of one function:
\begin{itemize}
  \item \texttt{get\_batch() -> (np.ndarray, np.ndarray)}

  This function retrieves the next batch of data in the form of $(\boldsymbol{x}, y)$.
\end{itemize}
The library provides a few standard wrappers for framework-specific data loaders, that we describe in the rest of this section.
All of them implement the \texttt{DataGenerator} interface.

When using this wrapper with the \texttt{fit\_generator} function of the TensorFlow or Keras classifier wrapper, training will be delegated to TensorFlow or Keras by actually calling its \texttt{fit\_generator} directly with the generator object.

\section{Versioning} \label{sec:versioning}
The library uses semantic versioning\footnote{https://semver.org}, meaning that version numbers take the form of MAJOR.MINOR.PATCH.
Given such a version number, we increment the
\begin{itemize}
  \item MAJOR version when we make incompatible API changes,
  \item MINOR version when we add functionality in a backwards-compatible manner, and
  \item PATCH version when we make backwards-compatible bug fixes.
\end{itemize}

Consistent benchmark results can be obtained with ART under constant MAJOR.MINOR versions.
Please report these when publishing experiments.

\subsection*{Acknowledgements}

We would like to thank the following colleagues (in alphabetic order) for their contributions, advice, feedback and support: Vijay Arya, Pin-Yu Chen, Evelyn Duesterwald, David Kung, Taesung Lee, Sameep Mehta, Anupama Murthi, Biplav Srivastava, Deepak Vijaykeerthy, Jialong Zhang, Vladimir Zolotov.

\bibliography{references}
\bibliographystyle{plainnat}

\end{document}